%% file: main.tex
\documentclass[10pt,twocolumn,letterpaper]{article}

\usepackage[pagenumbers]{cvpr} % To force page numbers, e.g. for an arXiv version

\usepackage{graphicx}
\usepackage{amsmath}
\usepackage{amssymb}
\usepackage{booktabs}
\usepackage{enumerate}
\usepackage{multirow}
\usepackage{graphicx}
\usepackage[normalem]{ulem}
\useunder{\uline}{\ul}{}
\usepackage{algorithm,algpseudocode}
\usepackage{amsfonts}
\usepackage{amssymb}
\usepackage{adjustbox}

\usepackage[pagebackref,breaklinks,colorlinks]{hyperref}

\usepackage[capitalize]{cleveref}
\crefname{section}{Sec.}{Secs.}
\Crefname{section}{Section}{Sections}
\Crefname{table}{Table}{Tables}
\crefname{table}{Tab.}{Tabs.}

\def\cvprPaperID{2908} % *** Enter the CVPR Paper ID here
\def\confName{CVPR}
\def\confYear{2022}

\newcommand{\Mc}[1]{\mathcal{#1}}

\input{shortcuts.tex}
\usepackage{etoolbox}
\patchcmd{\thanks}{#1}{\protect\doublespacing\normalsize#1}{}{}

\begin{document}

\title{Active Learning at the ImageNet Scale}

\author{Zeyad Ali Sami Emam\thanks{these authors contributed equally.}   \thanks{University of Maryland, College Park, MD.}   \thanks{National Institutes of Health, Bethesda, MD.}\\
{\tt\small zeyad@umd.edu}
\and
Hong-Min Chu\footnotemark[1] \footnotemark[2]\\
{\tt\small hmchu@umd.edu}
\and
Ping-Yeh Chiang\footnotemark[1] \footnotemark[2]\\
{\tt\small pchiang@umd.edu}
\and
Wojciech Czaja\footnotemark[2]\\
{\tt\small wojtek@umd.edu}
\and
Richard Leapman\footnotemark[3]\\
{\tt\small leapmanr@mail.nih.gov}
\and
Micah Goldblum\thanks{New York University, New York, NY.}\\
{\tt\small goldblum@nyu.edu}
\and
Tom Goldstein\footnotemark[2]\\
{\tt\small tomg@umd.edu}
}
\maketitle

%%%%%%%%% ABSTRACT
\begin{abstract}

    Active learning (AL) algorithms aim to identify an optimal subset of data for annotation, such that deep neural networks (DNN) can achieve better performance when trained on this labeled subset. AL is especially impactful in industrial scale settings where data labeling costs are high and practitioners use every tool at their disposal to improve model performance. The recent success of self-supervised pretraining (SSP) highlights the importance of harnessing abundant unlabeled data to boost model performance. By combining AL with SSP, we can make use of unlabeled data while simultaneously labeling and training on particularly informative samples.
    
    In this work, we study a combination of AL and SSP on ImageNet. We find that performance on small toy datasets -- the typical benchmark setting in the literature -- is not representative of performance on ImageNet due to the class imbalanced samples selected by an active learner. Among the existing baselines we test, popular AL algorithms across a variety of small and large scale settings fail to outperform random sampling. To remedy the class-imbalance problem, we propose Balanced Selection (BASE), a simple, scalable AL algorithm that outperforms random sampling consistently by selecting more balanced samples for annotation than existing methods. Our code is available at: \href{https://github.com/zeyademam/active\_learning}{https://github.com/zeyademam/active\_learning}.

\end{abstract}

%%%%%%%%% BODY TEXT
\section{Introduction}
\label{sec:intro}
    Fueled by the success of deep learning, the global data annotation market is projected to reach $\$3.4$ billion by 2028 \cite{Research2021global}. The data labeling process is a daunting hurdle for institutions aiming to deploy deep learning models at an industrial scale. The annotation process is slow, costly, and in some cases requires domain-expert annotators.
    
    A large body of machine learning research seeks to reduce data labeling costs by harnessing as much information as possible directly from unlabeled data or by leveraging other labeled datasets whenever possible. Ultimately, however, labeled data is required to achieve adequate deep learning model performance, especially in mission critical scenarios. Due to time and budget constraints, practitioners are often restricted to selecting a small subset of the available data for annotation. This restriction raises the following question: What is the best approach for selecting this subset?   
    
    Active learning (AL) is a subfield of machine learning (ML) dedicated to answering this question. Given a large pool of unlabeled data and a fixed labeling budget, an AL algorithm selects a subset of the unlabeled data to be annotated. Once labeled, the subset is subsequently used to train a ML model. The goal of the active learner is to select the subset that will optimize the generalization performance of the ML model.
    
    In this work, we focus on classification tasks using deep neural networks (DNN). We study a combination of AL and self-supervised pretraining (SSP) in the large data regime. Large-scale data is prevalent in real-world scenarios, where unlabeled data is typically abundant and cheap to collect. Furthermore, in real-world settings, practitioners are compelled to leverage the available unlabeled data in order to achieve adequate model performance at the lowest possible annotation cost. State-of-the-art SSP methods can provide these performance boosts at no annotation cost.  
    
    Prior research on AL focuses on the CIFAR-10, CIFAR-100, and SVHN \cite{Krizhevsky2009learning, Yang2019street} datasets to compare AL algorithms. However, it is unclear whether performance on these datasets is predictive of performance on real-world datasets that are orders of magnitude larger and that may contain many more classes or even imbalanced data. We focus particularly on ImageNet \cite{Russakovsky2015imagenet}, as it contains $1000$ classes, $1.2$ million images, and a significant amount of label noise \cite{Beyer2020imagenet, Stock2018convnets, Tsipras2020from} as is common in industrial settings \cite{Liao2021TowardsGP}. AL cost savings are much more impactful at the ImageNet scale and beyond, and cannot be understood by studying small datasets alone. These cost savings are due in part to the sheer amount of available data but also the ambiguity of the classes considered. To curate ImageNet, each image was presented to multiple human annotators who voted until a consensus was reached on the label \cite{Deng2009imagenet}. This voting mechanism translates directly to high annotation costs. 
        
    Finally, we specifically focus on the interaction of AL with SSP. SSP has been shown to provide a significantly larger accuracy boost compared to AL alone \cite{rethinkactivelearning}. It is therefore important to study whether AL offers any additional benefits on top of SSP. We present the first AL results on ImageNet using SSP.
    
    We summarize our contributions below:
    \begin{enumerate}
        \itemsep0em
        \item We demonstrate that the performance of popular AL methods, which has been observed on small datasets, does not transfer to the larger and more complex ImageNet challenge.  In fact, on the common linear evaluation task (see Section  \ref{sec:linear-eval-motivation}), most popular AL algorithms perform worse than random sampling on ImageNet.
        
        \item We identify and study sampling imbalance as a major failure mode for AL algorithms. Because ImageNet has many classes with highly heterogeneous properties, AL algorithms have a tendency to heavily sample from preferred classes while nearly ignoring others. This problem is less severe on simple tasks with fewer and more homogeneous classes.
        
        \item We introduce the Balanced Selection (BASE) AL strategy. BASE efficiently selects images that lie near class boundaries in feature space while also promoting an even class distribution. By carefully selecting which data to label, BASE achieves significantly better sample efficiency than standard self-supervised learning (SSL) pipelines that rely on random sampling.
        
        \item We show, for the first time, that AL can offer performance boosts on ImageNet when combined with SSL. Our BASE algorithm, when used to train a classifier on top of a SSL feature extractor, matches the top-5 accuracy of the state-of-the-art EsViT~\cite{esvit} SSL algorithm while using only $55\%$ of the labels.
    \end{enumerate}
    
    We make our code publicly available in the hopes that others can easily reproduce our results and use our codebase for future AL research\footnote{\href{https://github.com/zeyademam/active\_learning}{https://github.com/zeyademam/active\_learning}}. 
    %\footnote{\href{https://anonymous.4open.science/r/Active-Learning-At-The-ImageNet-Scale/README.md}{https://anonymous.4open.science/r/Active-Learning-At-The-ImageNet-Scale/README.md}}. 

%-------------------------------------------------------------------------
\section{Background \& Related Work}
    A typical AL algorithm cycles between learning from a small amount of labeled data, using the model to gather information about the unseen unlabeled data, and using this information to choose a subset of the unlabeled data to be manually annotated. This cycle then repeats, with the labeled dataset increasing in size at every iteration, until a pre-determined manual-annotation budget is exhausted. In this section, we provide a formal description of the AL problem, followed by an overview of existing methods. 
    
%-------------------------------------------------------------------------
%\subsection{Problem Statement}
    We will adopt the notation from \cite{sener2018active} with slight modifications to accommodate more general cases. We will study a $C$-way classification problem defined over a compact space $\Xcal=\RR^d$ and a finite label set $\Ycal = \{1, \ldots, C\}$. We denote the loss function by ${l(\cdot, \cdot, \textbf{w}): \Xcal, \Ycal \to \RR}$, where $\textbf{w}$ are the parameters (i.e., the weights) of a classifier $f(\textbf{w}, \cdot): \Xcal \to \Ycal$, which we simply denote as $f(x)$ in the rest of the paper. \\
    The entire dataset is a collection of $n$ points $Z\subseteq \Xcal \times \Ycal$ sampled \textit{i.i.d.} over $\Zcal = \Xcal \times \Ycal$ as $\{x_i, y_i\}_{i\in{[n]}} \sim p_\Zcal$. Initially, some subset of $m$ points is assumed to have been annotated by an expert, we will denote the indices of those points by $s^0 = \{s^0(j) \in [n]\}_{j\in [m]}$.\\
    An AL algorithm has access to $\{x_i\}_{i\in n}\subseteq \Xcal$ but only the labels with indices $s^0$, i.e. $\{y_{s^0(j)}\}_{j\in [m]}$. The algorithm is also given a budget $b$ of queries to ask an oracle (typically a human annotator), and a learning algorithm $A_s$ which outputs a set of parameters $\textbf{w}$ given $\{x_i\}_{i\in n}\subseteq \Xcal$ and $\{y_{s(j)}\}_{j\in [m]}$. The goal of AL is to identify a new subset $s^1$ of unlabeled data such that:
    
    \begin{equation}
        s^1 = \argmin_{s^1: |s^1|<b} \, {\EE_{x,y\sim p_\Zcal}} \left[l(x,y; A_{s^0\cup s^1}) \right]. \label{eq:al-loss}
    \end{equation}
    
    The above formulation constitutes one round of active learning. Typically, the algorithm runs for $K$ rounds producing a sequence of subsets $s^1, s^2 \ldots, s^{K}$ to be labeled by the oracle and added to the labeled dataset for the following round. We will denote by $D^k_L$ all the indices of points labeled before the start of round $k$, i.e., $D^k_L=\bigcup_{i=0}^{k-1} s_i$, and likewise $D_U^k=[n]\setminus D_L^k$ is the set of all unlabeled points at round $k$. We will use $|\cdot|$ to refer to the cardinality of a set. 
  
%\subsection{Existing AL Strategies}
    %\vspace{-2ex}
    \subsubsection*{Selection Methodologies}
        \label{sec:selection-methodology}
        
        Existing AL algorithms for DNNs can be roughly broken down into two categories: those designed to tackle class imbalanced datasets and those that are not. The wide majority of existing algorithms fall in the latter category.
        
        Algorithms designed for balanced datasets can be further broken down into two categories: uncertainty based sampling and density based sampling\cite{Aggarwal2020active}. Uncertainty based AL algorithms operate by first quantifying the classifier's uncertainty about its prediction on every unlabeled sample at round $k$ \cite{gal2015dropout, gal2017deep, Settles2010active}, then querying the examples on which the classifier is deemed most uncertain. Prediction entropy, least confidence, and margin sampling are commonly used uncertainty measures. Intuitively, uncertainty based sampling improves the model's prediction on subsets of the domain $\Xcal$ where the model is most uncertain, and in turn, improves the model's generalization ability.
        
        On the other hand, density based algorithms \cite{sener2018active, Ash2020deep} generate high-dimensional features from the data then select examples with the most representative features. In practice, this selection involves running a clustering algorithm \cite{Ash2020deep, Citovsky2021batch} or finding coresets \cite{sener2018active} in feature space. The features can be obtained by removing the network's linear classification head \cite{sener2018active, Citovsky2021batch} or by taking its gradients with respect to every sample \cite{Ash2020deep}. Intuitively, density based sampling ensures that the most densely populated regions of space, which contain the most data at test time, are represented in the labeled set.
        
        A smaller minority of AL algorithms specifically tackle class imbalanced data \cite{Aggarwal2020active}. However, as we discuss below, techniques in this area cannot scale to ImageNet.
        
    \subsubsection*{Scaling Ability}
        AL strategies designed for classical machine learning (ML) focus on querying a single label (i.e. $b=1$), re-training the model on all labeled data, querying the next example, retraining again, etc. This has obvious advantages as the active learner is given access to the current label and can therefore use it to guide its selection strategy. However, as datasets increased in size, ML algorithms became costly to train, and data annotation grew into an entire industry. It is now necessary for AL algorithms to operate at a large scale.
        
        To be practical for neural network applications, AL strategies must be able to query a large batch of data at once, receive labels for the entire batch, then query another batch (i.e., $b>>1$). This batch AL approach minimizes costs and time by keeping a large group of annotators occupied and by reducing the number of training runs needed to update DNNs on newly acquired data.
        At the same time, modern AL strategies must be able to efficiently process a massive pool of unlabeled data at each round (i.e., large $D_U^k$). 
        
        In Table \ref{tab:time_complexity} and Section \ref{sec:imagenet-scaling-issues}, we discuss the time complexities of several baselines considered in this paper in greater detail.
    
    \begin{table}[!htb]
        \centering
        \begin{tabular}{c|c}
            Algorithm & Time Complexity \\
            \hline
            Margin Sampler & $\Mc O(C \cdot \log(b) \cdot|D_U^k|)$\\
            Confidence Sampler & $\Mc O(C  \cdot \log(b) \cdot |D_U^k|)$\\
            Approx. Coreset \cite{sener2018active}  &  $\Mc O((b+|D_L^k|) \cdot d' \cdot |D_U^k|)$ \\
            BADGE \cite{Ash2020deep} & $\Mc O(C \cdot (b+|D_L^k|) \cdot d'' \cdot |D_U^k|)$ \\
            Balancing Sampler \cite{Aggarwal2020active} & $\Mc O(C \cdot b \cdot d' \cdot |D_U^k|)$\\
            BASE (ours) & $\Mc O(C\cdot (d' + \log(b)) \cdot |D_U^k|)$
        \end{tabular}
        \caption{Time complexities. $d'$ is the feature space dimension. On ImageNet using ResNet-50, $|D_U^k|\approx 10^6$, features $d' = 2048$, gradient embeddings $d''=2048 \times 10^3$, and $C=10^3$. In our experiments, $b=10^3$.}
        \label{tab:time_complexity}
        \vspace{-4mm}
    \end{table}
    
    \subsubsection*{Usage of Unlabeled Data}
        Unlabeled data is often abundant in real-world scenarios, and leveraging it effectively can lead to significant reductions in data annotation costs. In \cite{marginalactivelearning,rethinkactivelearning}, the authors study the benefits of AL when combined with both SSP and FixMatch, a semi-supervised technique. In this work, we restrict our experimental setup to SSP because semi-supervised techniques would require training to saturation multiple times on the entire Imagenet dataset \cite{sohn2020fixmatch}, which is prohibitively expensive. In Table \ref{tab:SSPisBetter}, we show that using only 25\% of the ImageNet labels, the MoCo v2 SSP method with randomly sampled labels is able to boost model performance by 18.15 percentage points, whereas carefully selecting examples using the VAAL \cite{vaal} AL algorithm only boosts performance by 1.5 percentage points if the classifier's weights are randomly initialized.  Clearly, initializing models with SSP offers strong advantages in the label scarce regime. In section \ref{sec:experiments}, we show for the first time that AL offers additional performance gains on top of SSP at the ImageNet scale.
        
        To the best of our knowledge, only two prior works \cite{Beluch2018thepower, vaal} study AL on ImageNet, and none study the interaction between AL and SSP at this scale. 
        
        \begin{table}[!htb]
            \centering
            \begin{tabular}{cccc}
            \toprule
                 \% of labels & SSP & Strategy & Accuracy \\
                 \midrule
                 \multirow{3}{*}{25\%} & No & Random & 50\%  \\
                 & No & VAAL & \begin{tabular}{c}
                 +1.5\% \end{tabular} \\
                 & Yes & Random & \begin{tabular}{c} +16.65\%  \end{tabular} \\
                 \bottomrule
            \end{tabular} 
            \caption{Performance gains offered by VAAL \cite{vaal} on ImageNet vs those offered by MoCo v2 \cite{Chen2020improved}, a popular SSP method. SSP, when combined with random sampling, yields a substantially larger boost in performance.} 
            \label{tab:SSPisBetter}
            \vspace{-4mm}
        \end{table}
       
%------------------------------------------------------------------------

\section{Linear Evaluation Task}
    \label{sec:linear-eval-motivation}
    When DNNs are deployed at an industrial scale, it is common to use a shared backbone network for feature extraction, then apply separate heads directly on the features to accomplish different downstream tasks. A single shared backbone is easier to maintain and carries a small memory footprint, making it more practical for edge devices. Training this entire pipeline end-to-end is time consuming and compute intensive. Therefore, in this setting, as new labeled data becomes available, the different task-specific heads are frequently finetuned while keeping the backbone frozen. The backbone itself is updated less frequently\footnote{The quintessential manifestation of this framework is described in Tesla's AI day: \href{https://youtu.be/j0z4FweCy4M?t=3300}{https://youtu.be/j0z4FweCy4M?t=3300} (55th minute).}. In fact, with the emergence of massive foundation models \cite{Bommasani2021opportunities}, such as GPT-3 \cite{Brown2020gpt3}, BERT \cite{Devlin2019bert}, and DALL-E \cite{Ramesh2021dalle}, practitioners may only have restricted access to the backbone. Consequently, it is important to evaluate AL algorithms when the feature extractor is fixed and only the classification head is finetuned. This task is a common benchmark in self-supervised learning (SSL) research \cite{esvit, Chen2020simple,Chen2020improved}, where the proposed SSL method is used to pretrain the network's feature extractor, then a linear classification head is trained on the features in a fully-supervised fashion. However, to the best of our knowledge, AL strategies on ImageNet have not been evaluated in this specific setting.
    
\section{Methods}
    \label{sec:our-method}
    We now describe our proposed method.  We start with a simple preliminary variant of uncertainty based selection.  We then show how this variant can be adapted to prevent sampling imbalances from emerging and accumulating over rounds, resulting in improved performance.
    \subsubsection*{Margin Selection}
        We first introduce a simplified variant of our Balanced Selection algorithm, which we call Margin Selection (MASE). MASE selects the $b$ examples closest to any decision boundary at every round of AL. Intuitively, these samples should have the most influence on the decision of the model. We define distance to decision boundary (DDB) as follows,
            \begin{align}
        \text{DDB} (x) = \min_{\epsilon} ||\epsilon||_2
       \quad \text{s.t.} f(x+\epsilon)  \neq f(x). 
        \end{align}
        When $f$ is a DNN, DDB is expensive to compute in input space. We instead estimate this distance in feature space. For the models considered in this paper, the final layer is a linear classification head on top of the features produced by a feature extractor; therefore, computing DDB in feature space reduces to computing the projection of the feature vector onto the normal vector of the linear decision boundary, which can be implemented very efficiently. We provide pseudocode for MASE in Algorithm \ref{alg:mase}.
        
        To the best of our knowledge, MASE is a novel AL strategy, similar algorithms were only studied on 2-class classification tasks using support vector machines \cite{Ertekin2007learning}, or a different definition of distance \cite{Cho2021least}. 
    
    \subsubsection*{Balanced Selection}
        In Section \ref{sec:experiments}, we show that only a single  baseline AL algorithms beats random sampling on the linear evaluation task on ImageNet and only by a relatively small margin. This is partially due to the imbalance induced by the active learner, which does not query examples evenly across classes. Motivated by this observation, we design Balanced Selection (BASE), an AL strategy capable of scaling efficiently while also querying a balanced batch of examples. As opposed to naively choosing the examples with the smallest DDB, BASE selects examples based on their distance to class specific decision boundaries (DCSDB), defined as
        \begin{align}
           &\text{DCSDB} (x, c) = \nonumber \\
           & \begin{cases}
           \min_{\epsilon} ||\epsilon||_2 \quad \text{s.t.} \;\; f(x+\epsilon)=c \;\; \text{if}\;\; f(x) \neq c  \\
           \min_{\epsilon} ||\epsilon||_2 \quad \text{s.t.} \;\; f(x+\epsilon)\neq c \;\;\text{if}\;\; f(x) = c.  \\
           \end{cases}
        \end{align}
        
\begin{figure}[!htb]
            \centering
            \includegraphics[width=1\linewidth]{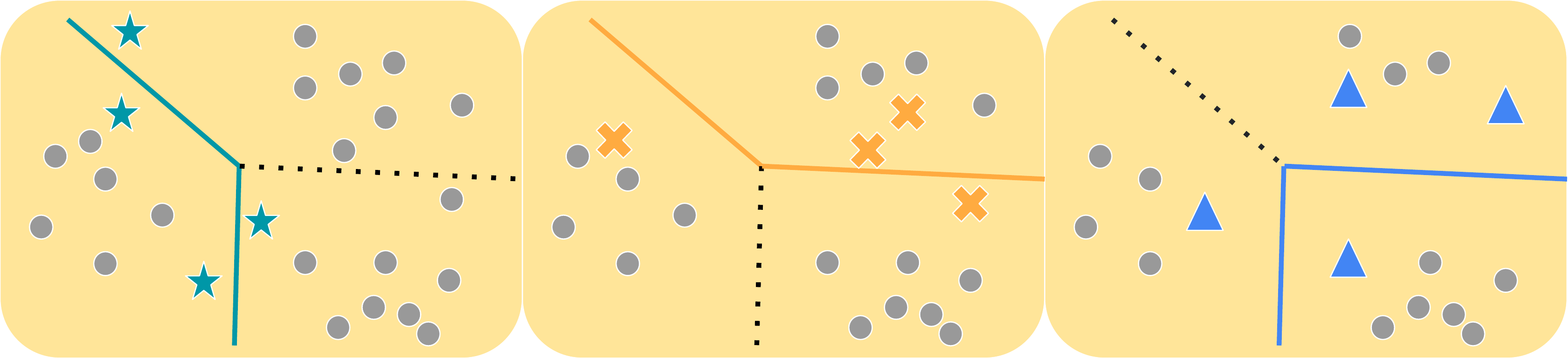}
            
            \caption{An illustration of BASE for 2-dimensional features and a 3 class problem. The algorithm selects an equal number of points (shown using colored stars, crosses, and triangles) that are closest to each decision boundary (solid lines).}
            \label{fig:method-illustration}
            \vspace{-4mm}
        \end{figure}

        More specifically, for each class $c\in \{1, \ldots, C\}$, BASE selects the $b/C$ samples with the smallest $\text{DCSDB} (x, c)$. Similar to our MASE implementation, we only consider distances in feature space. We provide a visual illustration of BASE's selection strategy in Fig.
        \ref{fig:method-illustration} and its pseudocode in Algorithm \ref{alg:base}. 
        The time complexity of Algorithm \ref{alg:base} is dominated by computing DCSDBs, but those can be computed once and stored, so the algorithm runs in $\Mc O(C\cdot (d' + \log(b)) \cdot |D_U^k|)$ in practice, where $d'$ is the dimension of the features. This puts BASE on par with the fastest baselines -- see \ref{tab:time_complexity}.
        
        \begin{algorithm}
        \caption{Algorithm for MASE}
        \begin{algorithmic}
            \State \textbf{Input:} Query Budget $b$, Indices of Labeled Samples $D_L^k$
            \State $s \gets D_L^k$
            \While{$|s|<|D_L^k|+b$}
                    \State $x^* \gets \argmin_{x \in [N] \backslash s} \text{DDB}(f(x))$ 
                    \State $s \gets s \cup \{x^*\}$
            \EndWhile
            \State \Return {$s \backslash D_L^k$}
        \end{algorithmic}
        \label{alg:mase}
        \end{algorithm}
        
%-------------------------------------------------------------------------

        \begin{algorithm}
        \caption{Algorithm for BASE}
        \begin{algorithmic}
            \State \textbf{Input:} Query Budget $b$, Number of Classes $C$, Indices of Labeled Samples $D_L^k$
            \State $s \gets D_L^k$
            \While{$|s|<|D_L^k|+b$}
                \For{$c$ in $[C]$}
                    \State $x^* \gets \argmin_{x \in [N] \backslash s} \text{DCSDB}(f(x)), c)$ 
                    \State $s \gets s \cup \{x^*\}$
                    \If{$|s|=|s^0|+b$}
                        \State \textbf{break}
                    \EndIf
                \EndFor
            \EndWhile
            \State \Return {$s \backslash D_L^k$}
        \end{algorithmic}
        \label{alg:base}
        \end{algorithm}
        
%-------------------------------------------------------------------------

%-------------------------------------------------------------------------
\section{Experiments}
    \label{sec:experiments}
    %\subsection{Experimental Setup}
        In this section, we outline our experimental design, followed by a presentation of our results.
        In figure captions, we will refer to different experiments using a capital letter for the dataset/model combination, and a capital roman numeral for the experimental setup. For example, setting \ref{setting:cifar10_resnet18}-\ref{setting:finetune_ssp} refers to AL strategies tested on CIFAR-10 using a ResNet-18 with the model weights initialized using SSP at every round. Baseline strategies are referenced using lower case roman numerals, e.g., \ref{baseline:BADGE} refers to BADGE. Below, we enumerate each dataset/model combination, training setup, and AL method, and assign each a letter or numeral. All other implementation details can be found in Appendix A.
        
    \subsubsection*{Datasets and Models}
        In our experiments, we use the following dataset and model architecture combinations. 
        \vspace{-1ex}
        \begin{enumerate}[A.]
            \itemsep0em 
            \item \label{setting:cifar10_resnet18} \textbf{CIFAR-10 \cite{Krizhevsky2009learning} w/ ResNet-18 \cite{He2016deep}.}
            \item \label{setting:imb_cifar10_resnet18} \textbf{Imbalanced CIFAR-10 \cite{Krizhevsky2009learning} w/ ResNet-18 \cite{He2016deep}:} The number of samples per class decreases exponentially from the most frequent class to the least frequent class; the most sampled class contains $10\times$ the number of samples in the least sampled class \cite{Cao2019ldam}. 
            \item \label{setting:imagenet_resnet50} \textbf{ImageNet \cite{Russakovsky2015imagenet} w/ ResNet-50 \cite{He2016deep}.}
            \item \label{setting:imagenet_vit} \textbf{ImageNet \cite{Russakovsky2015imagenet} w/ ViT \cite{Dosovitskiy2021an}.} 
        \end{enumerate}
     
    \subsubsection*{Training Setups} We consider two different settings for training the classifier.
        \vspace{-3ex}
        \begin{enumerate}[I.]
            \itemsep0em 
            \item \label{setting:finetune_ssp} \textbf{End-to-end finetuning from a self-supervised checkpoint.} We first train the network using self-supervised learning on all available unlabeled data. At every round $k$ of AL, the backbone's weights (all layers except the final linear classifier) are reset to the SSP weights then the network is finetuned end-to-end on all the available labeled data $D_L^k$.
            \item \label{setting:linear_eval} \textbf{Linear evaluation from a self-supervised checkpoint.} Here, we employ SSP. At every AL round $k$, we use the SSP checkpoint, but we only update the final linear layer of the network on labeled data $D_L^k$. 
        \end{enumerate}
        
    \subsubsection{Baselines} 
        We compare BASE (ours) to the following baselines. 
        \begin{enumerate}[i.]
            \itemsep0em 
            \item \label{baseline:RandomSampler}\textbf{Random Sampler.} Queries samples from $D_U^k$ uniformly at random. 
            \item \label{baseline:BalancedRandomSampler} \textbf{Balanced Random Sampler.} Iterates over classes and chooses an equal number of examples uniformly at random from each class. \textit{This baseline strategy cheats, as it requires the labels for points in $D_U^k$ in order to make its selection.  We include it only for scientific purposes.}
            \item \label{baseline:CoresetSampler} \textbf{Coreset AL. \cite{sener2018active}} We solve the k-center problem using the classical greedy 2-approximation. 
            \item \label{baseline:PartitionedCoreset} \textbf{Partitioned Coreset Sampler. \cite{Citovsky2021batch}} Partitions the dataset into p partitions, then runs the coreset algorithm to select  $b/p$ examples from each partition. This implementation only calculates pairwise distances on smaller subsets, which is more computationally efficient. 
            \item \label{baseline:BADGE} \textbf{BADGE AL \cite{Ash2020deep}.} Calculates the gradient with respect to the last linear layer, then applies the K-means++ seeding algorithm \cite{Arthur2007kmeans} on the gradients. On ImageNet, the size of the gradient embedding is proportional to the number of classes, which makes it $100 \times$ larger than CIFAR-10 embeddings. Furthermore, the K-means++ seeding algorithm requires the pairwise distances, which again is computationally prohibitive on ImageNet. 
            \item \label{baseline:PartitionedBADGE} \textbf{Partitioned BADGE Sampler. \cite{Citovsky2021batch}} BADGE with a similar partitioning trick as Partitioned Coreset, and global pooling to reduce the embedding dimension. 
            \item \label{baseline:ConfidenceSampler} \textbf{Confidence Sampler.} Selects the examples with the smallest top logit (least confidence). 
            \item \label{baseline:MarginSampler} \textbf{Margin Sampler\cite{minmargin}.} Selects examples with the smallest differences between the top logit and the second largest logit (minimum margin).
            \item \label{baseline:VAAL} \textbf{VAAL \cite{vaal}.} Trains a binary classifier to distinguish between features produced by labeled vs unlabeled samples. The features are obtained by training a variational autoencoder. Queries the unlabeled samples that the binary classifier is most confident about.
            \item \label{baseline:Balancing Sampler} \textbf{Balancing Sampler} \cite{Aggarwal2020active} Calculates cluster centers for each class in feature space, then targets the class with the least number of queried examples, and finally selects examples that are close to the target class' center and away from other clusters. 
            \item \label{baseline:MarginDistanceSampler} \textbf{MASE (ours).} See Section \ref{sec:our-method}.  
        \end{enumerate}
        
%-------------------------------------------------------------------------
% CIFAR-10 plots

    \begin{figure}[!htb]
      \centering
        \includegraphics[width=0.8\linewidth]{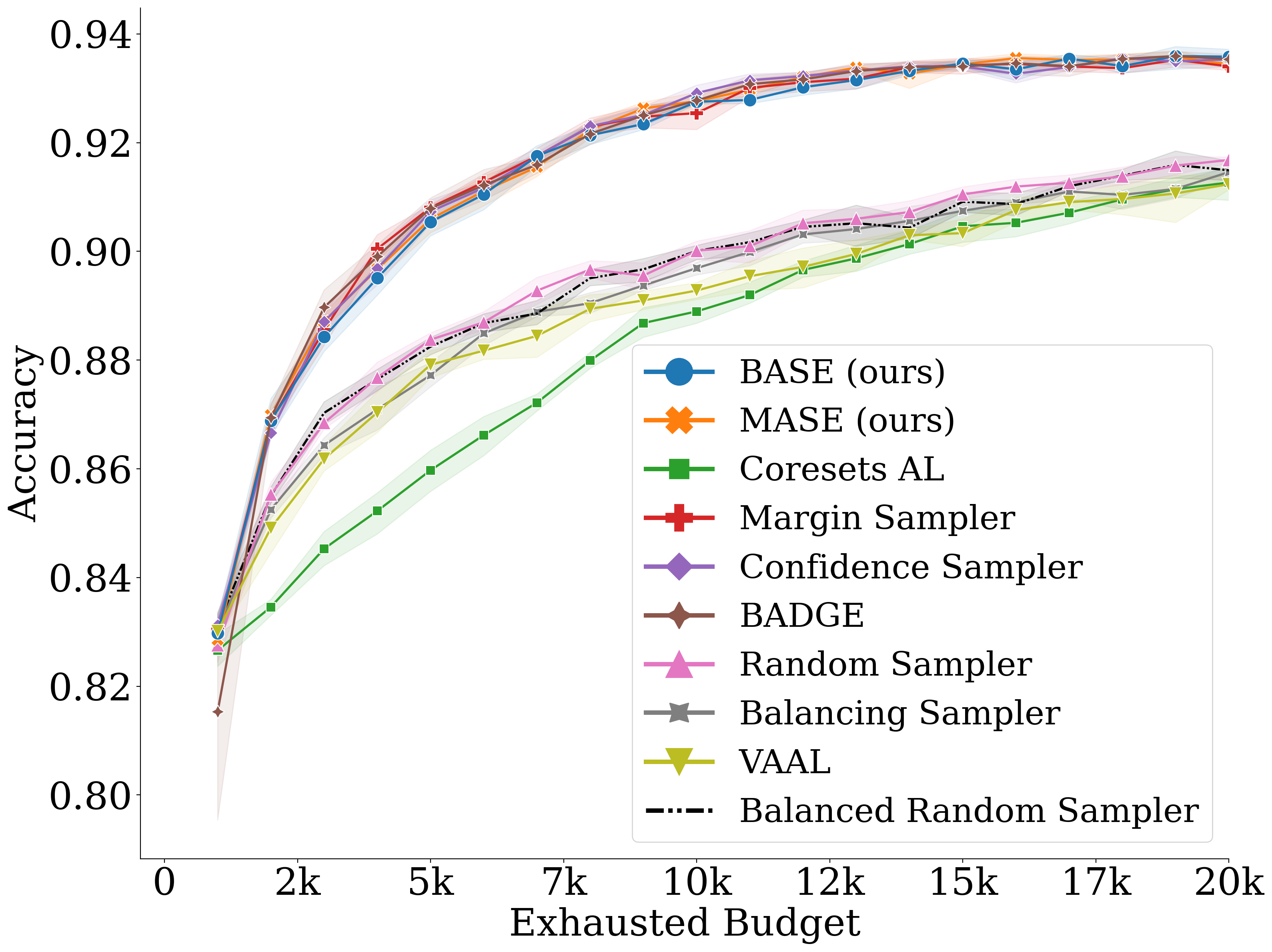}
        \caption{Setting \ref{setting:cifar10_resnet18}-\ref{setting:finetune_ssp} with $|s^0|=b=1000$. Average results over 5 runs on CIFAR-10 obtained by training a ResNet-18 end-to-end starting from a SSP checkpoint at every AL round. Shaded regions depict the $95\%$ confidence interval of the results. }
        \label{fig:cifar10_test_acc_ip1k_b1k}
        \vspace{-3ex}
    \end{figure}

%-------------------------------------------------------------------------
% IMAGENET PLOTS

    \begin{figure*}[!htb]
        \centering
        % Accuracy Plots 
        \begin{subfigure}{0.48\linewidth}
            \includegraphics[width=0.9\linewidth]{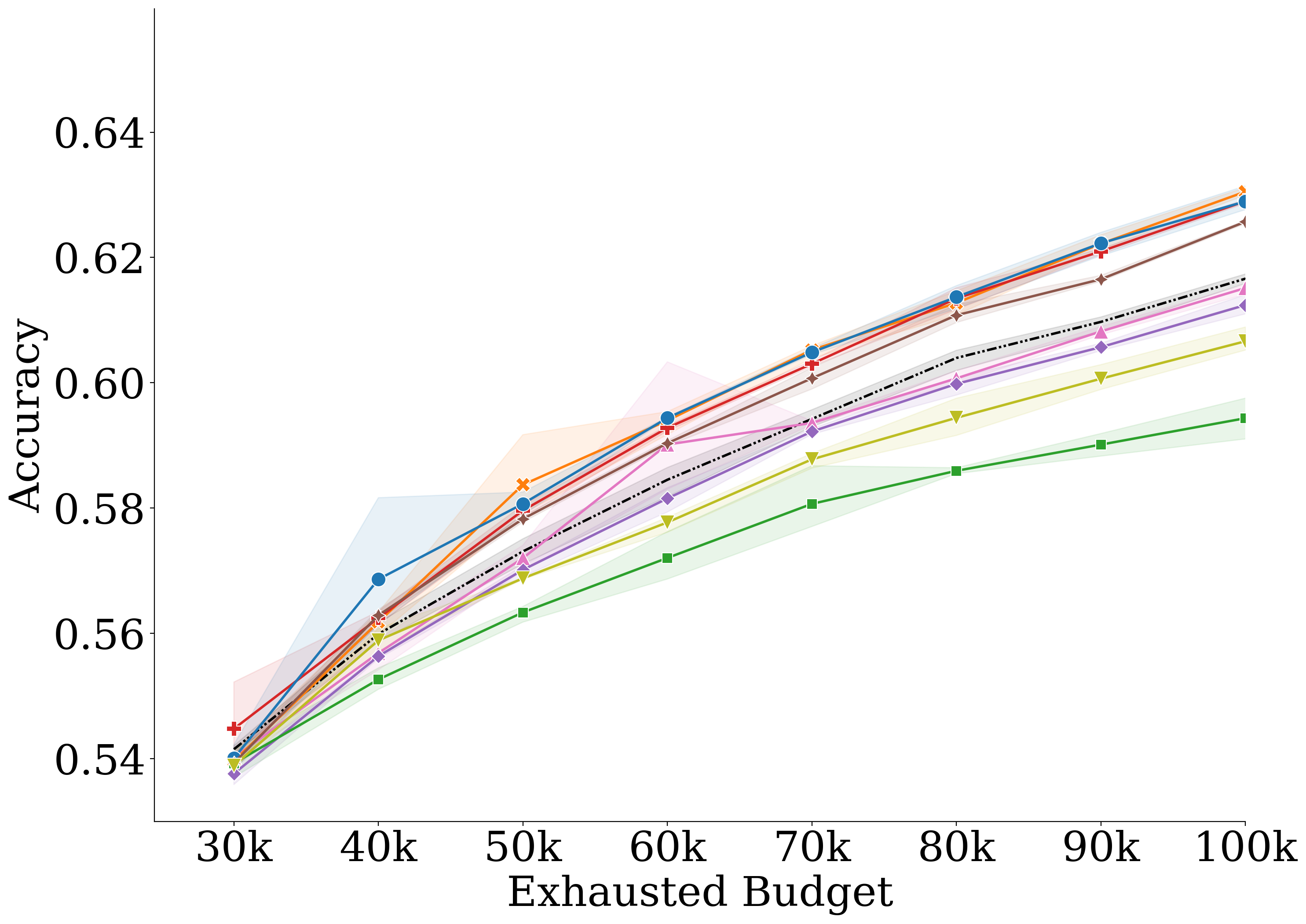}
            \caption{Setting \ref{setting:imagenet_resnet50}-\ref{setting:finetune_ssp}}
            \label{fig:imagenet_finetune_test_acc}
        \end{subfigure}
        \hfill
        \begin{subfigure}{0.48\linewidth}
            \includegraphics[width=0.9\linewidth]{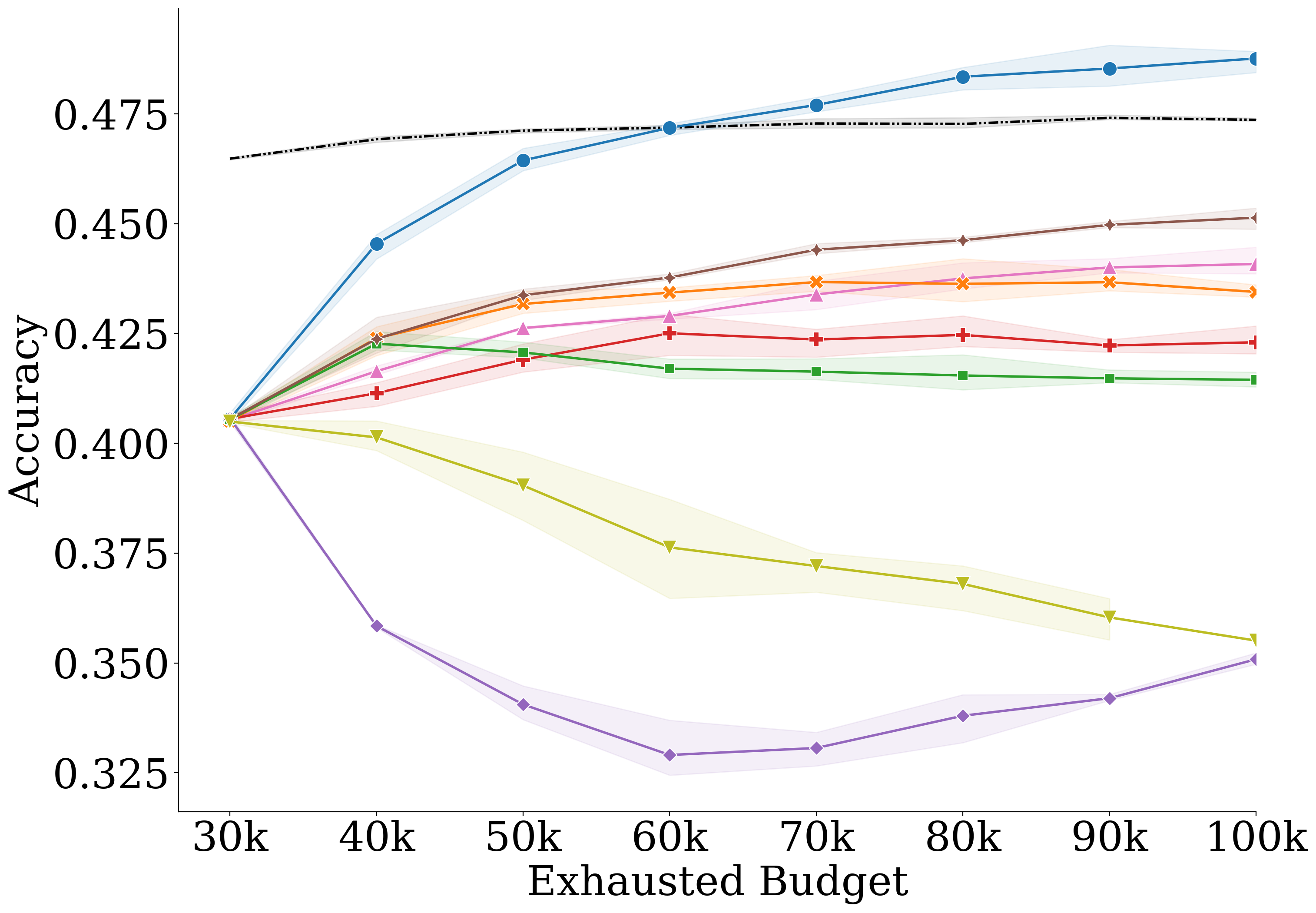}
            \caption{Setting \ref{setting:imagenet_resnet50}-\ref{setting:linear_eval}}
            \label{fig:imagenet_linear_test_acc}
        \end{subfigure}
    
        \bigskip
        
        \centering
        \begin{subfigure}{0.96\linewidth}
        \includegraphics[trim={0 8cm 0 10cm},clip, width=0.9\linewidth]{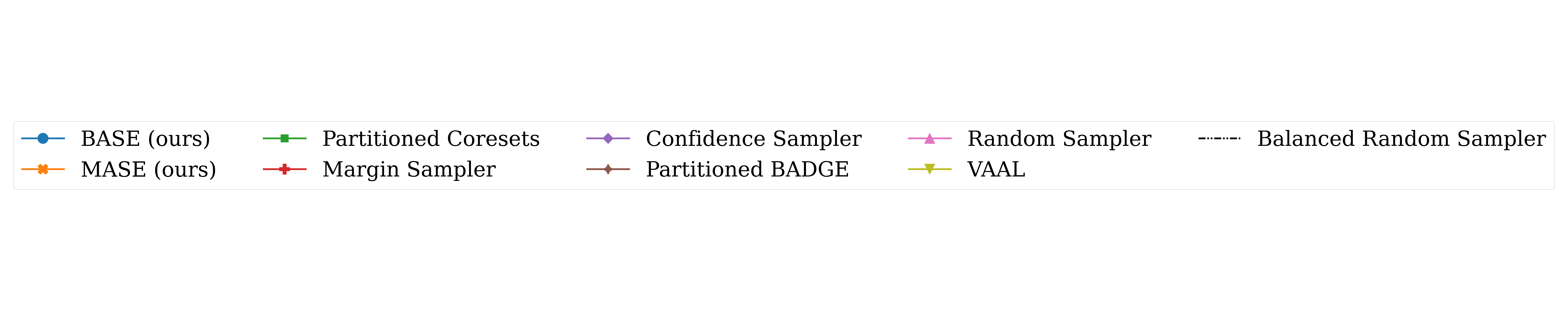}
        \end{subfigure}
        
        \bigskip
        % Imb Ratio Plots 
        \centering
        \begin{subfigure}{0.48\linewidth}
            \includegraphics[width=0.9\linewidth]{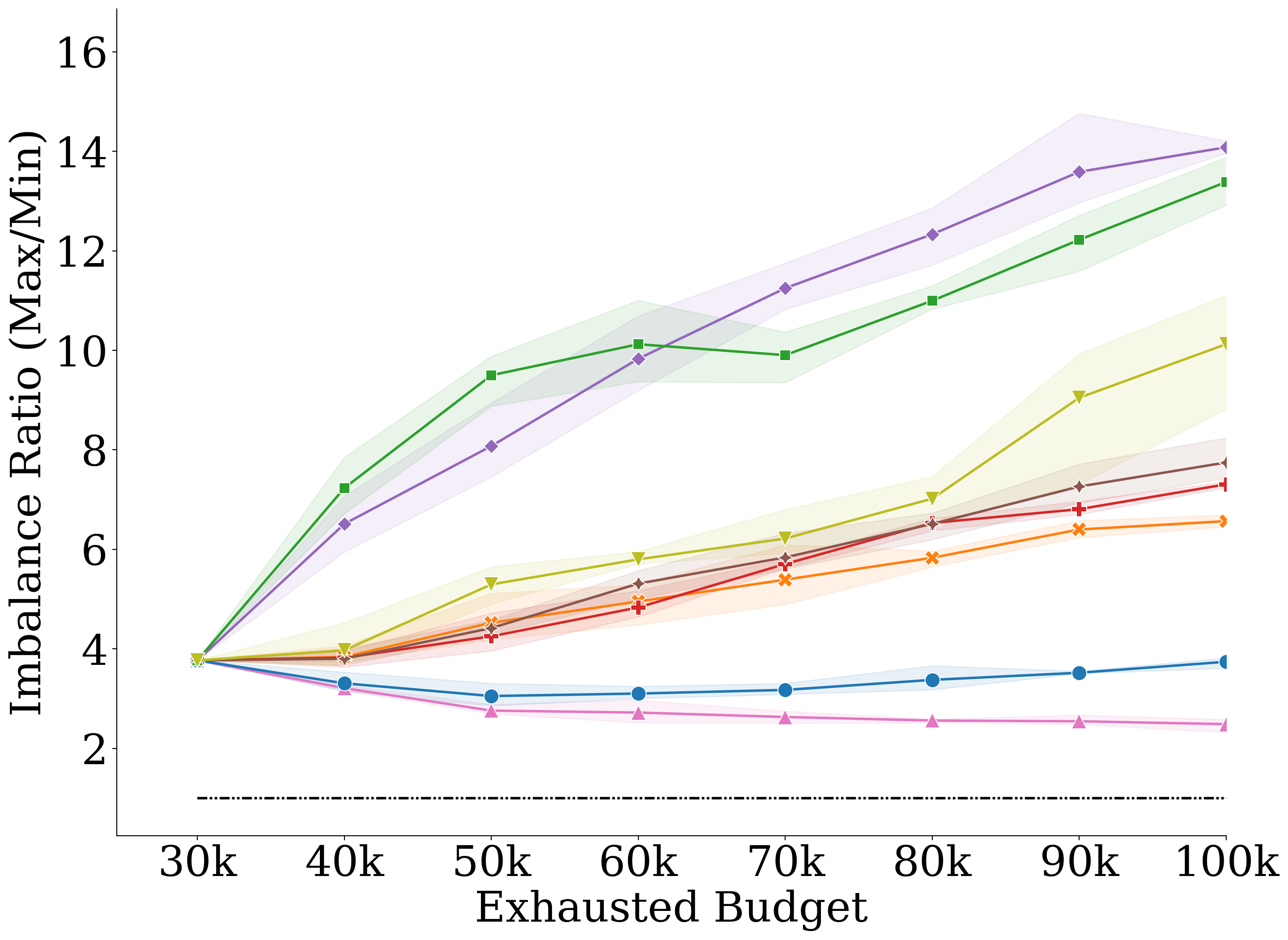}
            \caption{Setting \ref{setting:imagenet_resnet50}-\ref{setting:finetune_ssp}}
            \label{fig:imagenet_finetune_imb_ratio}
        \end{subfigure}
        \hfill
        \begin{subfigure}{0.48\linewidth}
            \includegraphics[width=0.9\linewidth]{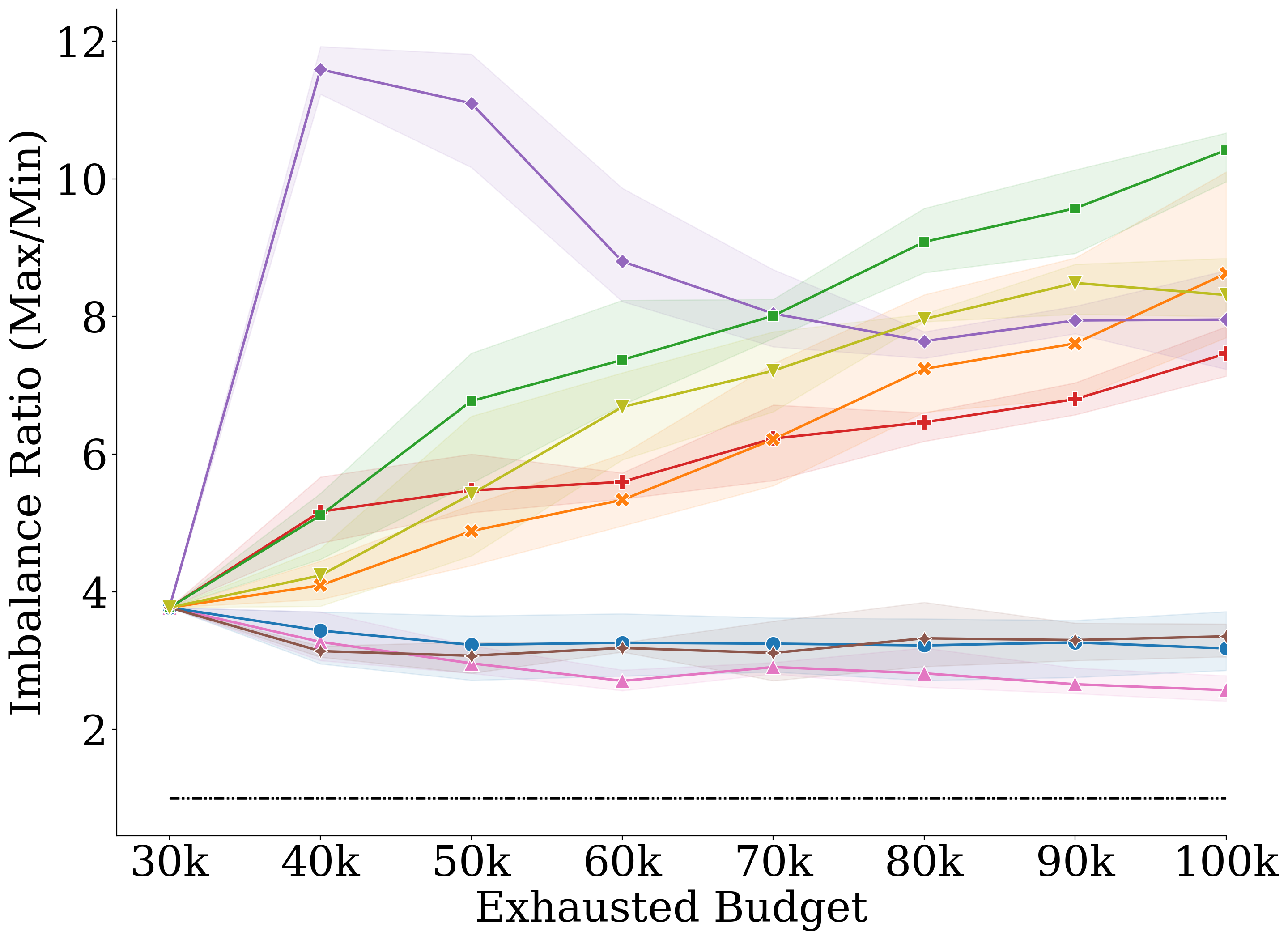}
            \caption{Setting \ref{setting:imagenet_resnet50}-\ref{setting:linear_eval}}
            \label{fig:imagenet_linear_imb_ratio}
        \end{subfigure}
        \caption{Average ImageNet accuracy and imbalance ratio over 3 runs. Shaded regions depict the $95\%$ confidence interval of the results. Figures \ref{fig:imagenet_finetune_test_acc} and \ref{fig:imagenet_finetune_imb_ratio} are obtained by by training a ResNet-50 end-to-end starting from a SSP checkpoint at every AL round. In contrast, Figures \ref{fig:imagenet_linear_test_acc} and \ref{fig:imagenet_linear_imb_ratio} are obtained by finetuning only the final linear layer at every round.} %Imbalance ratios in figures \ref{fig:imagenet_finetune_imb_ratio} and \ref{fig:imagenet_linear_imb_ratio} are computed by taking the ratio of the number of queried samples belonging to the most sampled class and that of the least queried class at each round of AL.}
    \end{figure*}
  
% LINEAR HISTOGRAMS 
    \begin{figure*}[!htb]
        \centering
        \begin{subfigure}{.01\linewidth}
            \centering
            \includegraphics[height=14.85\linewidth ]{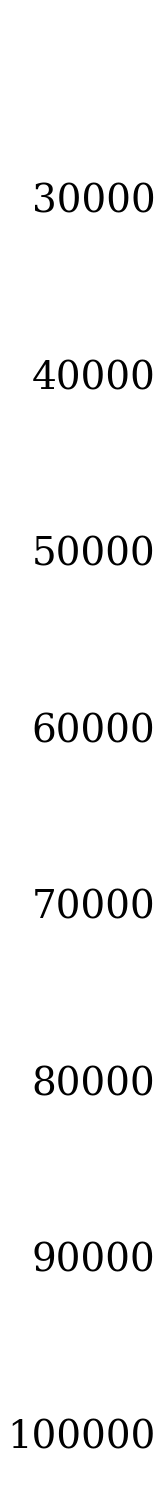}
            \caption*{\phantom{hi}}
        \end{subfigure}
        \hspace{.2mm}
        \begin{subfigure}{.11\linewidth}
            \centering
            \includegraphics[width=0.9\linewidth]{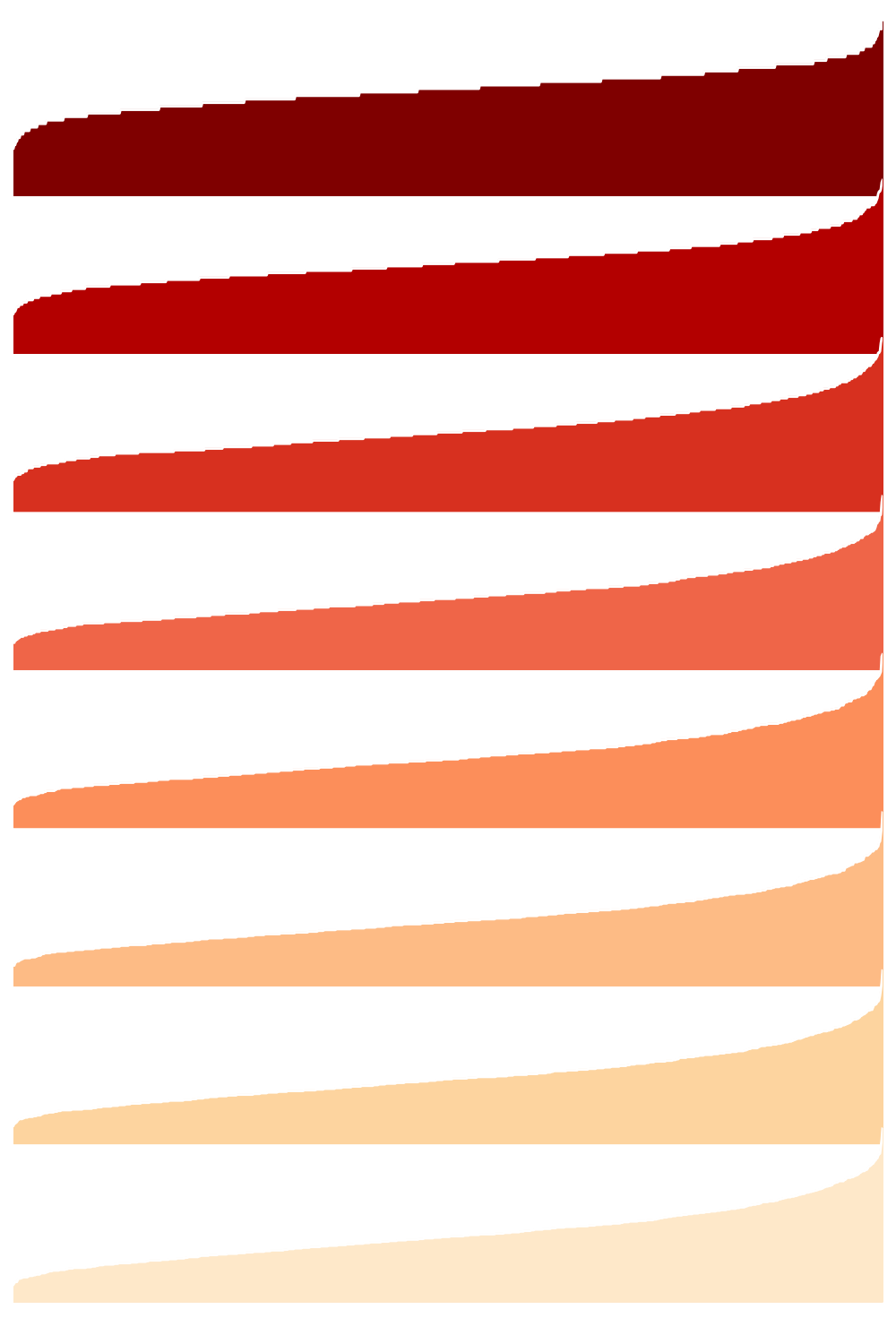}
            \caption{\ref{baseline:PartitionedCoreset}}
        \end{subfigure}
            \hfill
        \begin{subfigure}{.11\linewidth}
            \centering
            \includegraphics[width=0.9\linewidth]{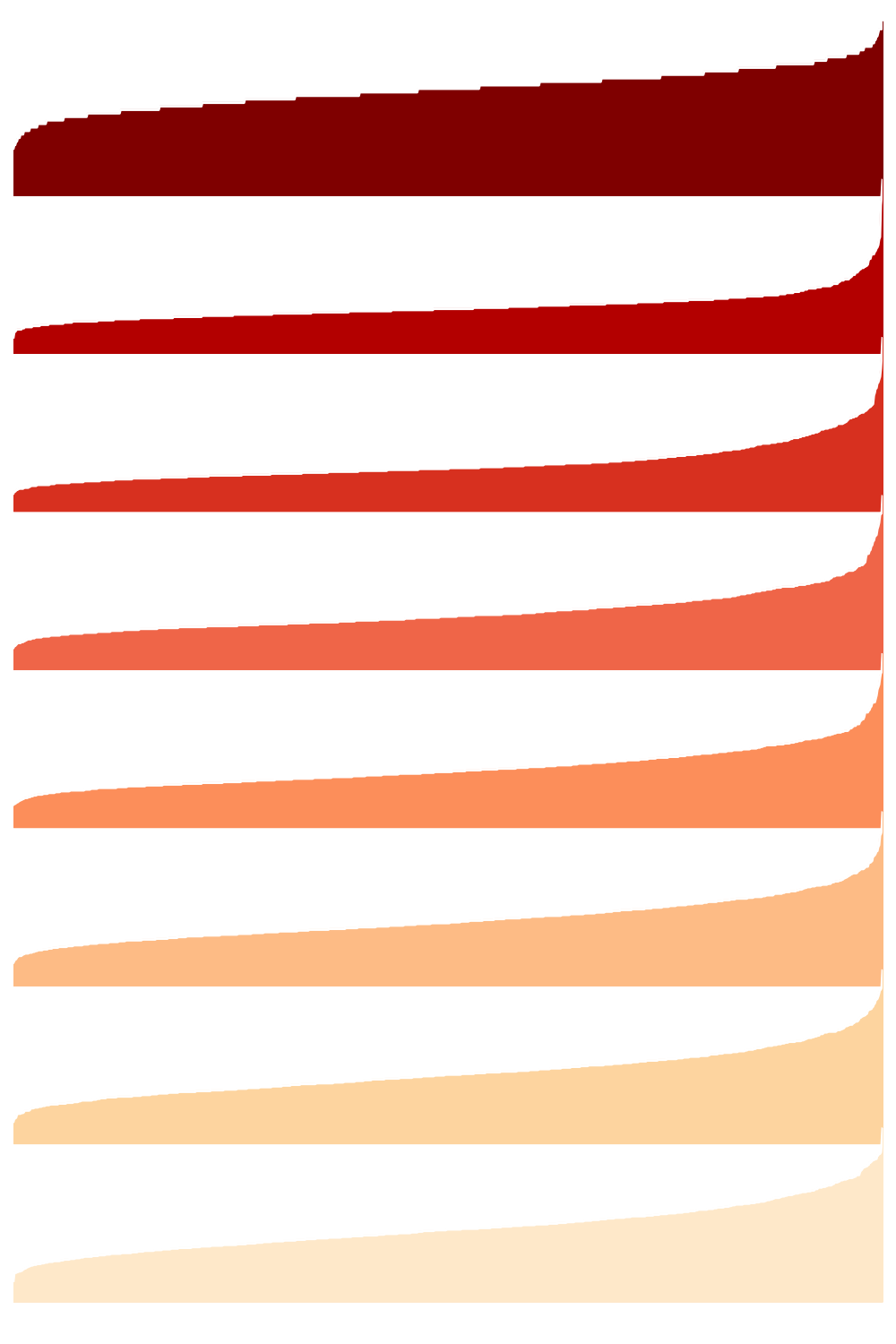}
            \caption{\ref{baseline:ConfidenceSampler}}
        \end{subfigure}
         \hfill
        \begin{subfigure}{.11\linewidth}
            \centering
            \includegraphics[width=0.9\linewidth]{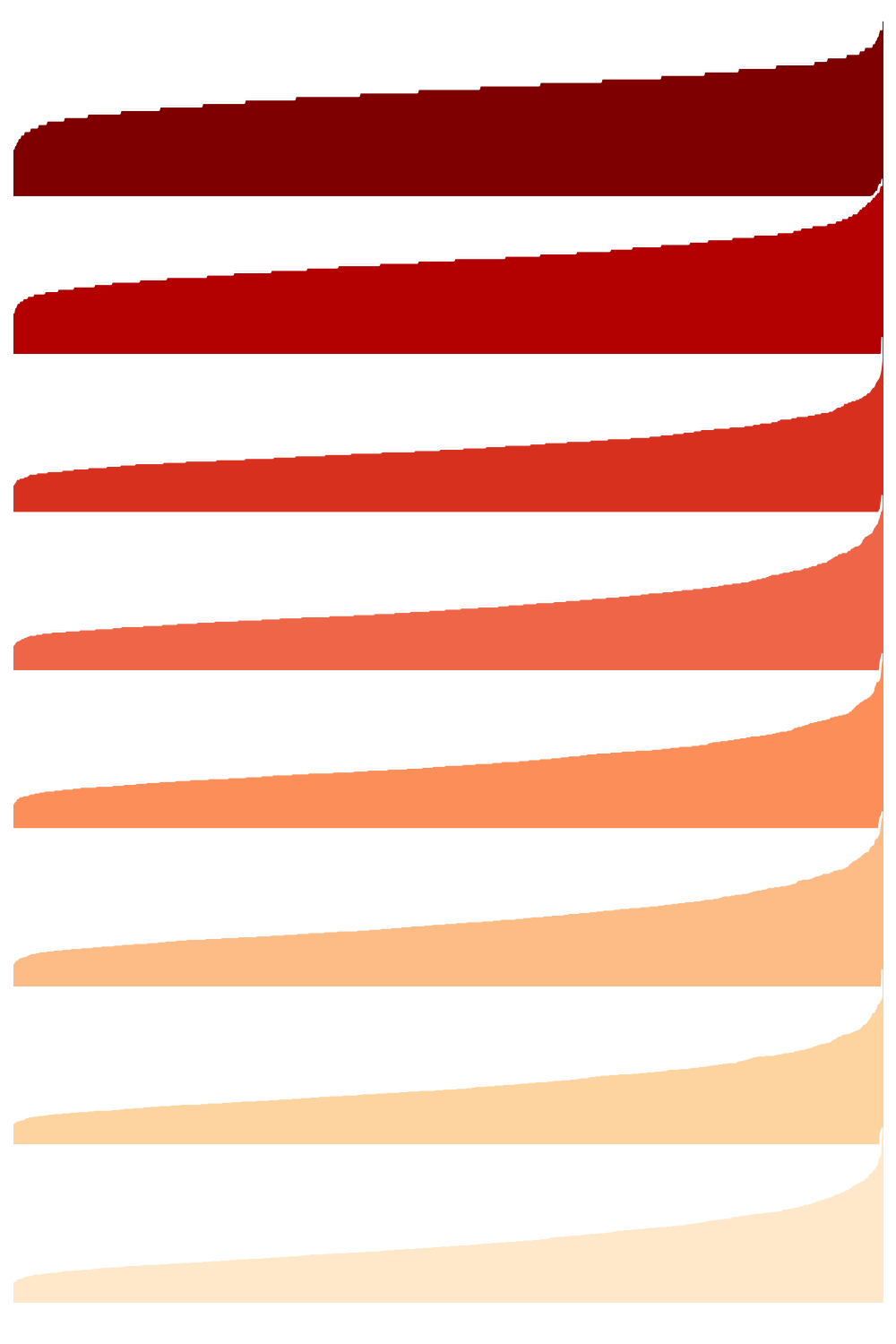}
            \caption{\ref{baseline:VAAL}}
        \end{subfigure}
         \hfill
        \begin{subfigure}{.11\linewidth}
            \centering
            \includegraphics[width=0.9\linewidth]{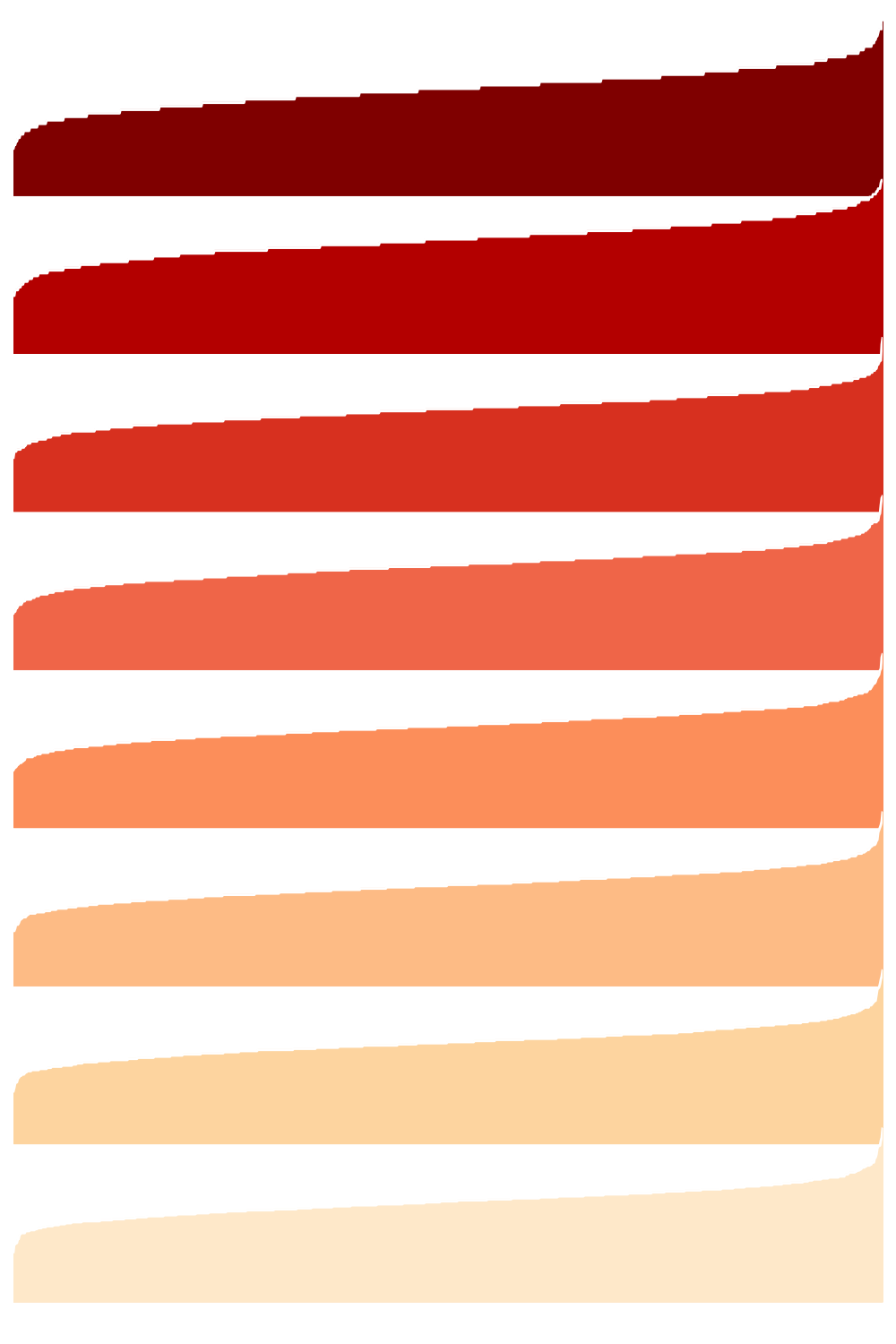}
            \caption{\ref{baseline:PartitionedBADGE}}
        \end{subfigure}
            \hfill
        \begin{subfigure}{.11\linewidth}
          \centering
          \includegraphics[width=0.9\linewidth]{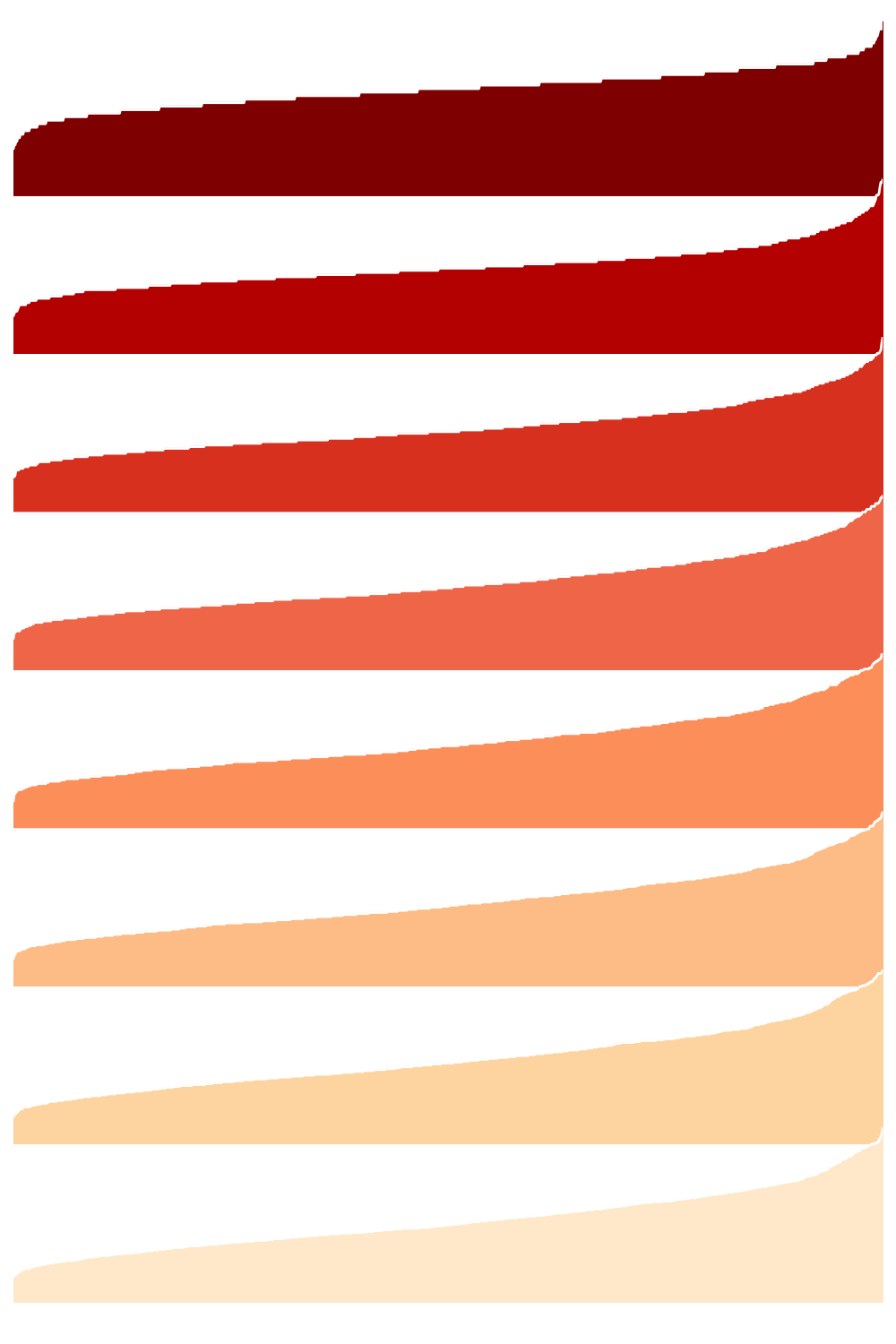}
          \caption{\ref{baseline:MarginSampler}}
        \end{subfigure} 
           \hfill
        \begin{subfigure}{.11\linewidth}
            \centering
            \includegraphics[width=0.9\linewidth]{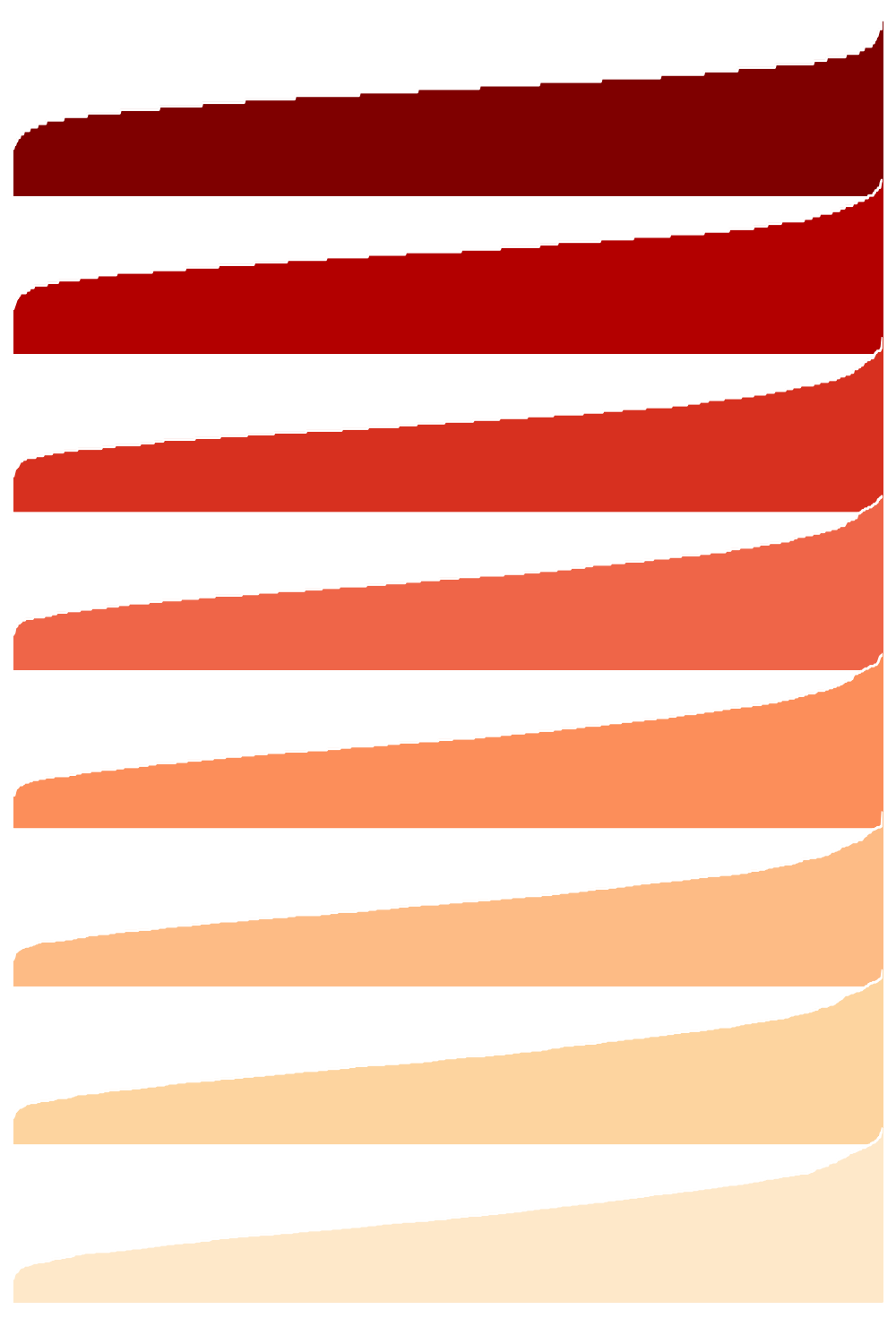}
            \caption{MASE \ref{baseline:MarginDistanceSampler}}
        \end{subfigure}
            \hfill
        \begin{subfigure}{.11\linewidth}
            \centering
            \includegraphics[width=0.9\linewidth]{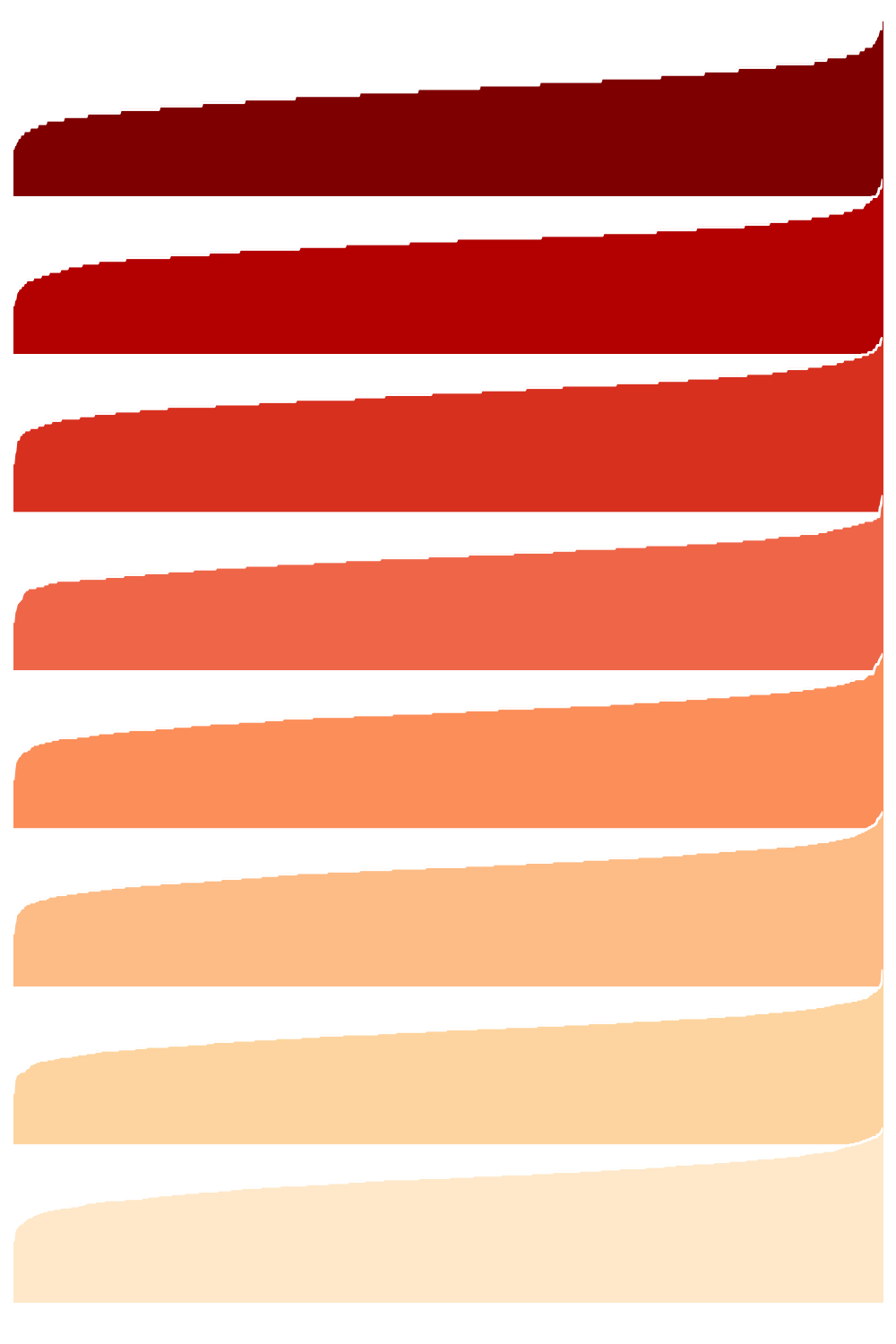}
            \caption{BASE (ours)}
        \end{subfigure}
            \hfill
        \begin{subfigure}{.11\linewidth}
            \centering
            \includegraphics[width=0.9\linewidth]{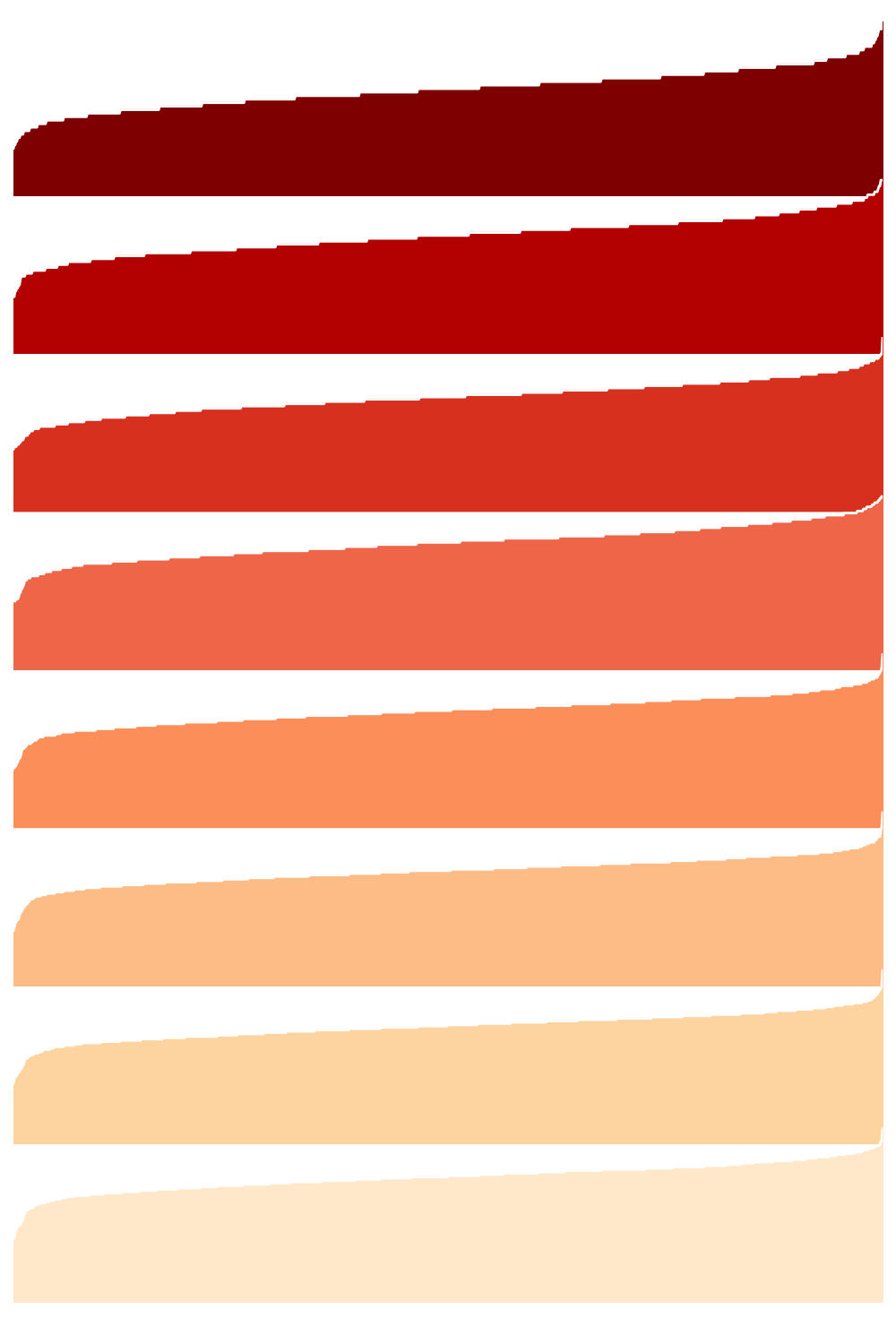}
            \caption{\ref{baseline:RandomSampler}}
        \end{subfigure}
        \caption{Setting \ref{setting:imagenet_resnet50}-\ref{setting:linear_eval}. The distribution of $D_U^k$ at every AL round for different strategies on ImageNet in the linear evaluation setting. All experiments start with the same randomly selected subset $s^0$. The x-axis is sorted for each histogram (every row in every subplot) from least queried class to most queried class. The height of the histogram at a given location on the x-axis indicates the proportion of the examples sampled from that class. BASE is visibly the most balanced strategy after random sampling.}
        \label{fig:imagenet_linear_histograms}
        \vspace{-3ex}
    \end{figure*}
%-------------------------------------------------------------------------

% IMBALANCED CIFAR-10 plots
    \begin{figure*}[!htb]
        \centering
        \begin{subfigure}{0.48\linewidth}
            \includegraphics[width=0.9\linewidth]{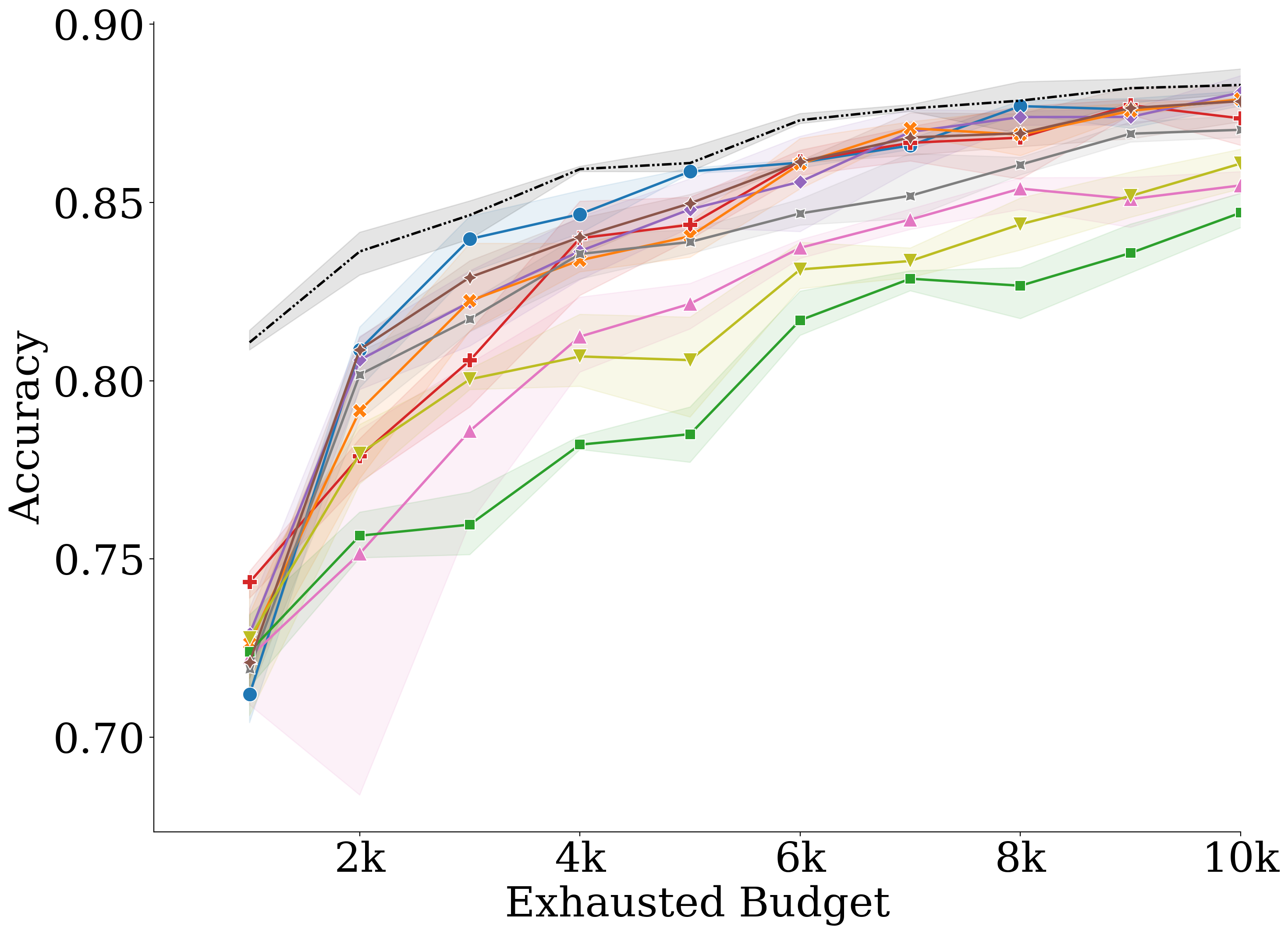}
            \caption{\phantom{a}}
        \end{subfigure}
        \hfill
        \begin{subfigure}{0.48\linewidth}
            \includegraphics[width=0.9\linewidth]{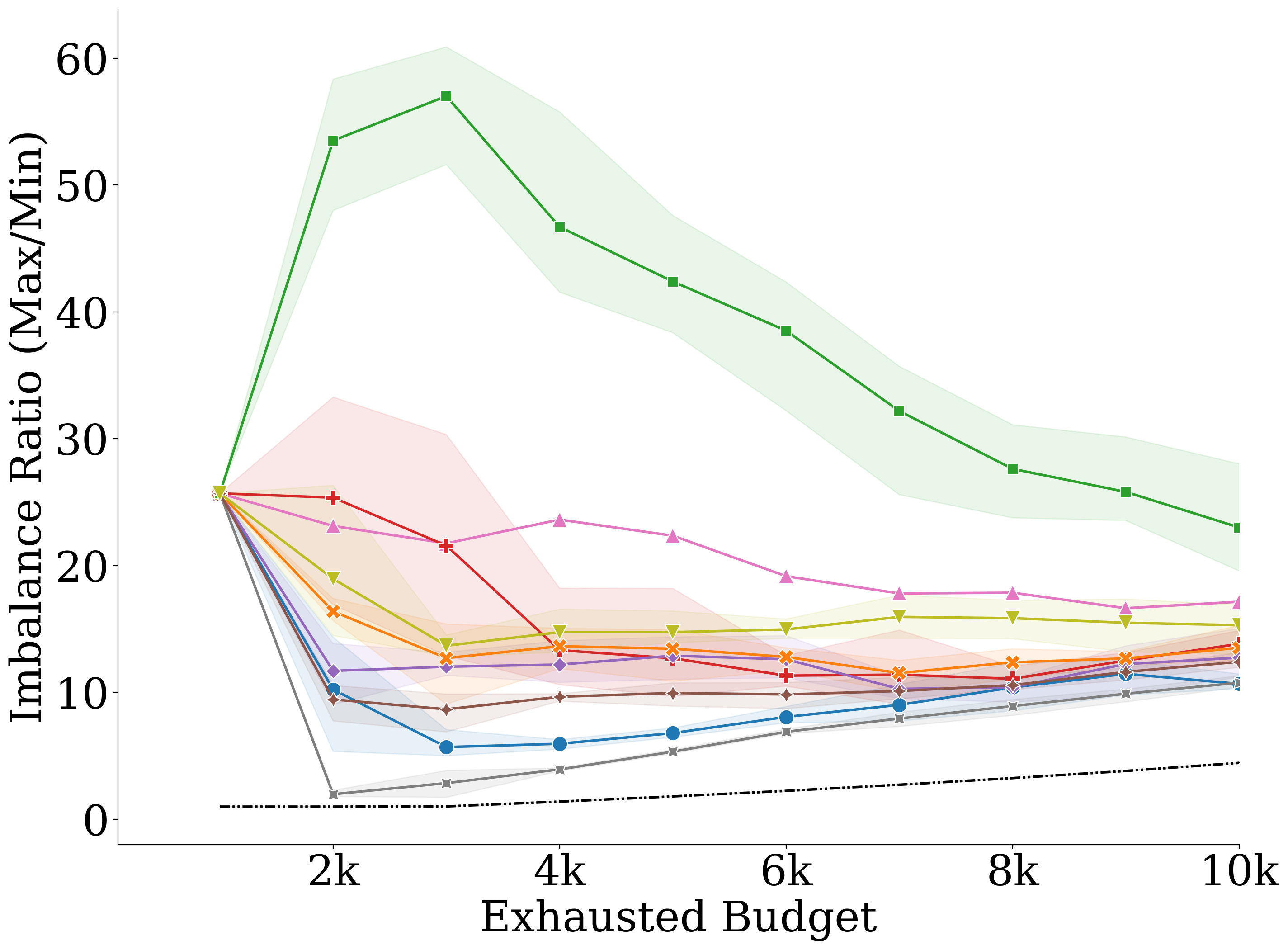}
            \caption{\phantom{b}}
            \label{fig:imbcifar10_imb_ratio}
        \end{subfigure}

        \bigskip
        
        \centering
        \begin{subfigure}{0.96\linewidth}
            \includegraphics[trim={0 8cm 0 10cm},clip, width=0.9\linewidth]{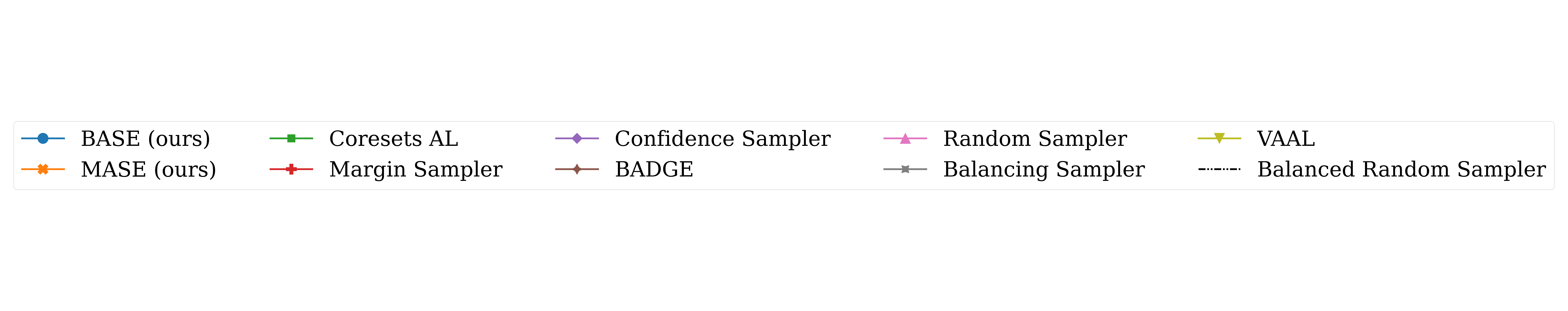}
        \end{subfigure}
        
        \caption{Setting \ref{setting:imb_cifar10_resnet18}-\ref{setting:finetune_ssp} with $|s^0|=b=1000$. Average results over 3 runs on imbalanced CIFAR-10 obtained by finetuning a ResNet-18 end-to-end starting from a SSP checkpoint at every AL round. Shaded regions depict the $95\%$ confidence interval of the results.} 
        \label{fig:imbcifar10}
    \end{figure*}
    
%-------------------------------------------------------------------------
% ESVIT TABLE

    \begin{table*}[!htb]
    \centering
        \begin{tabular}{ll|cccccc}
        \toprule
                                        &          &                      \multicolumn{5}{c}{ Number of Labels (\% of all labels)}                                                                                \\ 
                                        & Strategy & 
                                             0.3M (24\%)   & 
                                             0.5M (39\%)  & 
                                             0.7M (55\%)  & 
                                             0.9M (71\%)  & 
                                             1.1M (87\%)  &  
                                             All(100\%)          \\ \midrule
        \multirow{2}{*}{Top 1 Acc.} & BASE (ours)                                          & 78.9\% & 80.5\% & 81.0\%                & {\ul \textbf{81.2\%}} & 81.2\%                & 81.2\%                \\
                                        & Random                                        & 78.9\% & 79.9\% & 80.5\%                & 80.7\%                & 81.0\%                & {\ul \textbf{81.2\%}} \\
        \multirow{2}{*}{Top 5 Acc.} & BASE (ours)                                         & 94.4\% & 95.2\% & {\ul \textbf{95.5\%}} & 95.5\%                & 95.5\%                & 95.5\%                \\
                                        & Random                                         & 94.5\% & 94.9\% & 95.1\%                & 95.3\%                & {\ul \textbf{95.5\%}} & 95.5\%          \\    \bottomrule 
                                        
        \end{tabular}
        \caption{When applying BASE on the linear evaluation task with the state-of-the-art EsViT \cite{esvit} SSP method, we are able to achieve the same state-of-the-art linear evaluation accuracy on ImageNet with only 71\% (for top-1 acc.) and 55\% (for top-5 acc.) of the data. The bolded number indicates the smallest amount of required data to achieve the same accuracy as using all of the data for training.}
        \label{tab:esvit}
        \vspace{-3mm}
    \end{table*}

%-------------------------------------------------------------------------

    \subsubsection{Scalability of baseline methods}
        \label{sec:imagenet-scaling-issues}
        Coreset (\ref{baseline:CoresetSampler}) and BADGE (\ref{baseline:BADGE}) are prohibitively expensive to run at the ImageNet scale -- see Table \ref{tab:time_complexity}. Additionally, both algorithms require storing large tensors in memory -- $O(d' \cdot |D_U^k|)$ space complexity -- as they require solving an optimization problem in feature space. For this reason, we exclude these methods from our comparisons, and instead implement Partitioned Coreset Sampler (\ref{baseline:PartitionedCoreset}) and Partitioned BADGE (\ref{baseline:PartitionedBADGE}), two scalable variants of the original strategies \cite{Citovsky2021batch}.
        Finally, we highlight that the Balancing Sampler (\ref{baseline:Balancing Sampler}) is not a batch AL algorithm as it acquires labels one at a time. This restriction makes the algorithm completely impractical in terms of both computation and human labeling bandwidth in large-scale settings.

    \subsection{Solving Class Imbalance Allows Scaling}
        This section will show that class balanced sampling is important to scaling. Without explicitly imposing balance, imbalance becomes a problem and causes baseline algorithms to under-perform random sampling at the ImageNet scale. But by querying balanced samples, our BASE algorithm recovers the good properties of AL in the large-scale regime, and even matches the state-of-the-art EvSiT \cite{esvit} results using only 71\% of the ImageNet labels.
    
        \subsubsection{Baselines Perform Poorly on ImageNet}
            Here, we analyze the performance of all baselines on ImageNet. In Figure \ref{fig:imagenet_finetune_test_acc}, we compare different baselines on ImageNet, starting from a SSP checkpoint and finetuning the network end-to-end at each AL round. Three baselines provide material performance boosts over random sampling in that setting: Margin Sampler (\ref{baseline:MarginSampler}), Partitioned BADGE (\ref{baseline:PartitionedBADGE}), and our MASE algorithm (\ref{baseline:MarginDistanceSampler}).
            In Figure \ref{fig:imagenet_linear_test_acc}, we evaluate all baselines on ImageNet in the linear evaluation setting described in Section \ref{sec:linear-eval-motivation}. Surprisingly, \textit{only a single baseline outperforms random sampling on the linear evaluation task.}
        \subsubsection{The Importance of Balanced Sampling}
            Figures \ref{fig:imagenet_finetune_imb_ratio} and \ref{fig:imagenet_linear_imb_ratio} compare {\em class imbalance ratios} -- the number of labels from the most sampled class over that of the least sampled class -- for each sampling strategy. Most baseline samplers disproportionately query certain classes. Indeed, the Confidence Sampler, the worst performing baseline in Figure \ref{fig:imagenet_linear_test_acc}, induces an imbalance ratio close to 12 after the first round of AL. To further investigate the effects of class imbalance, we implement a cheating baseline strategy (Balanced Random Sampler \ref{baseline:BalancedRandomSampler}) which queries a perfectly balanced batch at each round. On the end-to-end finetuning experiment in Figure \ref{fig:imagenet_finetune_test_acc}, querying balanced batches is not sufficient to outperform random sampling. However, on the linear evaluation task in Figure \ref{fig:imagenet_linear_test_acc}, the cheating Balanced Random Sampler (\ref{baseline:BalancedRandomSampler}) outperforms random sampling by approximately 6 percentage points at every round.
            
            Class imbalance ratios do not fully describe the class imbalance across all classes, but only the extremes. To further investigate class imbalance, we analyze the distributions of $D_L^k$ for all baselines on the ImageNet linear evaluation task in Figure \ref{fig:imagenet_linear_histograms}. It is clear from the figure that all baselines exhibit long tailed distributions, and increase the imbalance over time.
            In Appendix B, we include yet another measure of class imbalance using entropy.
            
        \subsubsection{BASE Outperforms Baselines and Mitigates Class Imbalance}
            In Section \ref{sec:our-method}, we proposed BASE, an AL algorithm specifically designed to query class balanced data. BASE can significantly outperform random sampling on the ImageNet linear evaluation task shown in Figure \ref{fig:imagenet_linear_test_acc}. In fact, BASE can even outperform the unrealistic Balanced Random Sampler (\ref{baseline:BalancedRandomSampler}), which cheats by using knowledge of ground truth labels to achieve perfect class balance. On the finetuning experiment in Figure \ref{fig:imagenet_finetune_test_acc}, BASE performs on par with the best two baselines in terms of accuracy. However, in Figures \ref{fig:imagenet_linear_histograms}, \ref{fig:imagenet_finetune_imb_ratio}, and \ref{fig:imagenet_linear_imb_ratio}, we show that our AL algorithm consistently achieves a more uniform class distribution than all other baselines on both the linear evaluation and end-to-end finetuning ImageNet tasks. The class distribution histograms for end-to-end finetuning can be found in Appendix C.
            
            Finally, in Table \ref{tab:esvit},  we show that by carefully selecting examples, BASE can reproduce the state-of-the-art linear evaluation top-1 accuracy results reported in EsViT \cite{esvit} using $\thicksim 29\%$ less labeled data; BASE only needs $55\%$ of the labels to match the same top-5 accuracy reported in \cite{esvit}.
        
    \subsection{Class Imbalance on Small Datasets}
        Motivated by our observations concerning the importance of balanced sampling in large-scale settings, we also investigate whether the balancing aspect of BASE offers benefits in small-scale settings. To this end, in Figure \ref{fig:imbcifar10}, we compare all AL strategies on an imbalanced version of CIFAR-10~\cite{imbalance} with an imbalance ratio of $10$ starting from a SSP checkpoint obtained by training on the full imbalanced dataset. In this experiment, when training the classifier, we weigh each class differently in the loss function to penalize rare classes more heavily. The plots show a strong correlation between class distributions and  performance of the AL algorithm. Balanced Random Sampler (\ref{baseline:BalancedRandomSampler}), the ``cheating" algorithm, achieves the best accuracy across the board. And with minor exceptions, for each algorithm, the better the performance in terms of accuracy, the less severe the observed class imbalance.
        
        BASE is the best performing strategy in terms of accuracy and only second best in terms of class imbalance -- the best being the non-scalable Balancing Sampler (\ref{baseline:Balancing Sampler}). 
       
    \subsection{AL Performs Differently with SSP}
        Throughout this paper, we argue that applying SSP along with simple random sampling is much more powerful than applying AL alone -- see Table \ref{tab:SSPisBetter}. Therefore, AL algorithms must prove that they can outperform random sampling in the SSP setting, otherwise they are redundant and potentially harmful to performance. In Figure \ref{fig:cifar10_test_acc_ip1k_b1k}, we show that some popular baselines, notably, Coreset AL (\ref{baseline:CoresetSampler}), are indeed harmful. A potential explanation for this failure mode can be found in \cite{Anonymous2022best}, where the authors show that warm-starting the network weights at each round can negatively impact the performance of Coreset AL (\ref{baseline:CoresetSampler}). Warm-starting means to continue training the network starting with the weights obtained in the previous AL round, as opposed to randomly re-initializing the network weights at every round (cold start). We suspect that SSP, just like warm-starting, may negatively impact the performance of Coreset AL on CIFAR-10 and conclude that future research should not draw conclusions about the performance of an AL algorithm in the SSP setting solely by observing its behaviour in the cold starting setting.

%-------------------------------------------------------------------------
\vspace{-1ex}
\section{Conclusion}
    AL for DNNs is a very difficult problem to study, partly because we still cannot answer very fundamental questions about the generalization abilities of DNNs \cite{Huang2020understanding}, but also because \textit{random sampling is an incredibly robust baseline}. In this paper, we highlighted the importance of stress-testing AL algorithms where they are most useful, namely on large-scale tasks. We showed that popular existing works cannot compete with random sampling across all settings, and we designed BASE, a robust AL strategy capable of doing just that. In future work, we hope to tackle more complex problems, where the cost savings incurred by AL are even more dramatic, such as large-scale segmentation and detection tasks. 

%%%%%%%%% REFERENCES
{\small
\bibliographystyle{ieee_fullname}
\bibliography{egbib}
}

\newpage
\appendix
%--------------------------------------------APPENDIX ---------------------------------------%

\section{Additional Experimental Details}

We provide additional details of implementations and hyperparameters in the following sections.

\subsection{Dataset Details and Early Stopping}

\textbf{Dataset Division} We split the target dataset into training set, validation set and test set.
All AL algorithms are restricted to query from the training set, and the initial pool is also sampled from training set. The validation set is used for early stopping.

CIFAR-10 comes with natural split of training and testing data. 
We keep the testing data as test set. We randomly sample $1\%$ of the training data as validation set and keep the rest as training set.

For Imbalanced CIFAR-10, we keep the original testing split as test data. We then follow \href{https://github.com/kaidic/LDAM-DRW/blob/master/imbalance_cifar.py}{this implementation} 
to subsample a set of long-tailed imbalance data from the training split. 
The set of imbalance data is then randomly partitioned into training/validation data with $0.99/0.01$ split ratio.

For ImageNet, the dataset itself comes with natural training split and validation split.
We use the validation split as test set. We randomly sample $10\%$ of the training split as validation set and keep the rest as training set.

\textbf{Early Stopping.} We use the validation set to estimate the final test accuracy and perform early stopping during the network training to avoid over-fitting. In particular, we stop training the classification network if the validation performance stops improving after a specific number of rounds (specified as a hyperparameter).

\subsection{Hyperparameters for Each Experiment Setting}

We conduct our experiments in the following four settings: \ref{setting:imagenet_resnet50}-\ref{setting:finetune_ssp}, \ref{setting:imagenet_resnet50}-\ref{setting:linear_eval}, \ref{setting:cifar10_resnet18}-\ref{setting:finetune_ssp}, and \ref{setting:imb_cifar10_resnet18}-\ref{setting:finetune_ssp}.

We provide the hyperparameters shared across these settings in Table~\ref{tab:hyper_param}, 
and discuss setting specific hyperparameters as follows.
For setting \ref{setting:cifar10_resnet18}-\ref{setting:finetune_ssp}, and \ref{setting:imb_cifar10_resnet18}-\ref{setting:finetune_ssp}, we use cosine annealing learning rate scheduler with  $T_{max} = 200$.
For setting \ref{setting:imagenet_resnet50}-\ref{setting:finetune_ssp} and \ref{setting:imagenet_resnet50}-\ref{setting:linear_eval}, we start from the learning rate provided in Table~\ref{tab:hyper_param}, and decrease it by a factor of $0.1$ every 20 epochs.

\subsection{Additional Details on baselines}
In this section we discuss additional implementation details of Partitioned Coreset/BADGE sampler and VAAL sampler.

\textbf{Partitioned Coreset/BADGE Sampler.}
To enable Coreset and BADGE sampler to run on ImageNet, we modify each algorithm to allow to scale, inspired by the approach in \cite{Citovsky2021batch}.
At each AL round we partition each of $D_U^k$ and $D_L^k$ into 10 random partitions. We then take one partition from each and combine them into 10 partitions, say $P_1, \ldots, P_{10}$, then run coresets or BADGE separately on each partition using $b/10$ budget. 

For BADGE we use global average pooling on the gradient embeddings to reduce their dimension to 512.

\textbf{VAAL Sampler.}
We follow the \href{https://github.com/sinhasam/vaal}{repository} provided by the original paper~\cite{vaal} to implement VAAL sampler. 
Since the architecture of the Variational Auto-Encoder (VAE) provided in the repository fails to handle ImageNet naturally, 
we instead use the VAE architecture in~\cite{Tolstikhin2017wae} and follow the paper to calculate the unsupervised loss with 
randomly-cropped $64\times 64$ patches instead of the full original images.
Also, we perform a single VAE optimizer step for every classifier optimizer step. The original \cite{vaal} paper does not comment on this, however, their codebase performs two VAE optimizer steps for every classifier optimizer step. 

\section{Distribution Analysis with Entropy}
\begin{figure*}
    \centering
    \begin{subfigure}{.48\linewidth}
        \centering
        \includegraphics[width=.98\linewidth]{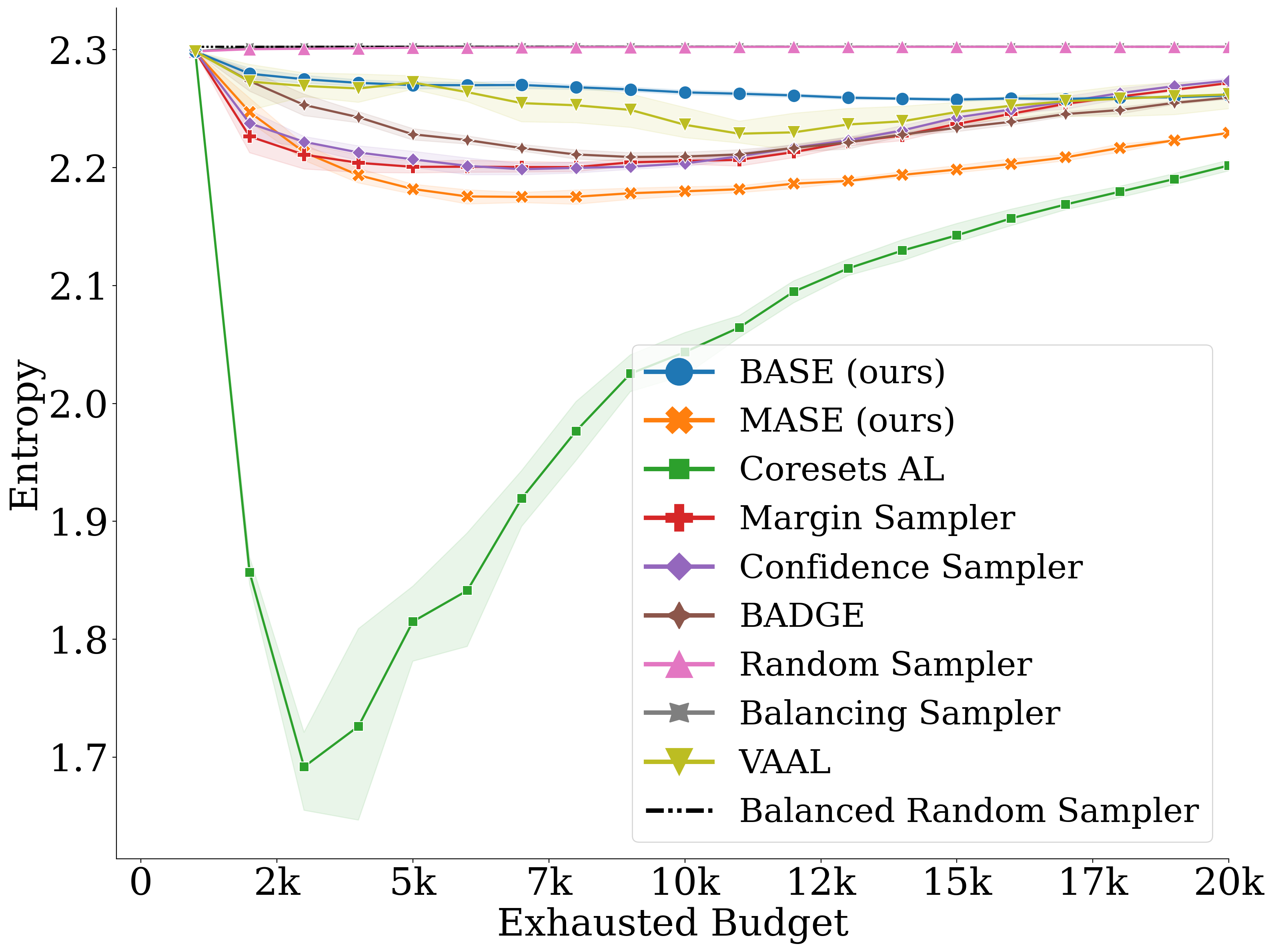}
        \caption{Setting \ref{setting:cifar10_resnet18}-\ref{setting:finetune_ssp}. }
        \label{fig:entropy_cifar10_finetune}
    \end{subfigure}
    \hfill
    \begin{subfigure}{.48\linewidth}
        \centering
        \includegraphics[width=.98\linewidth]{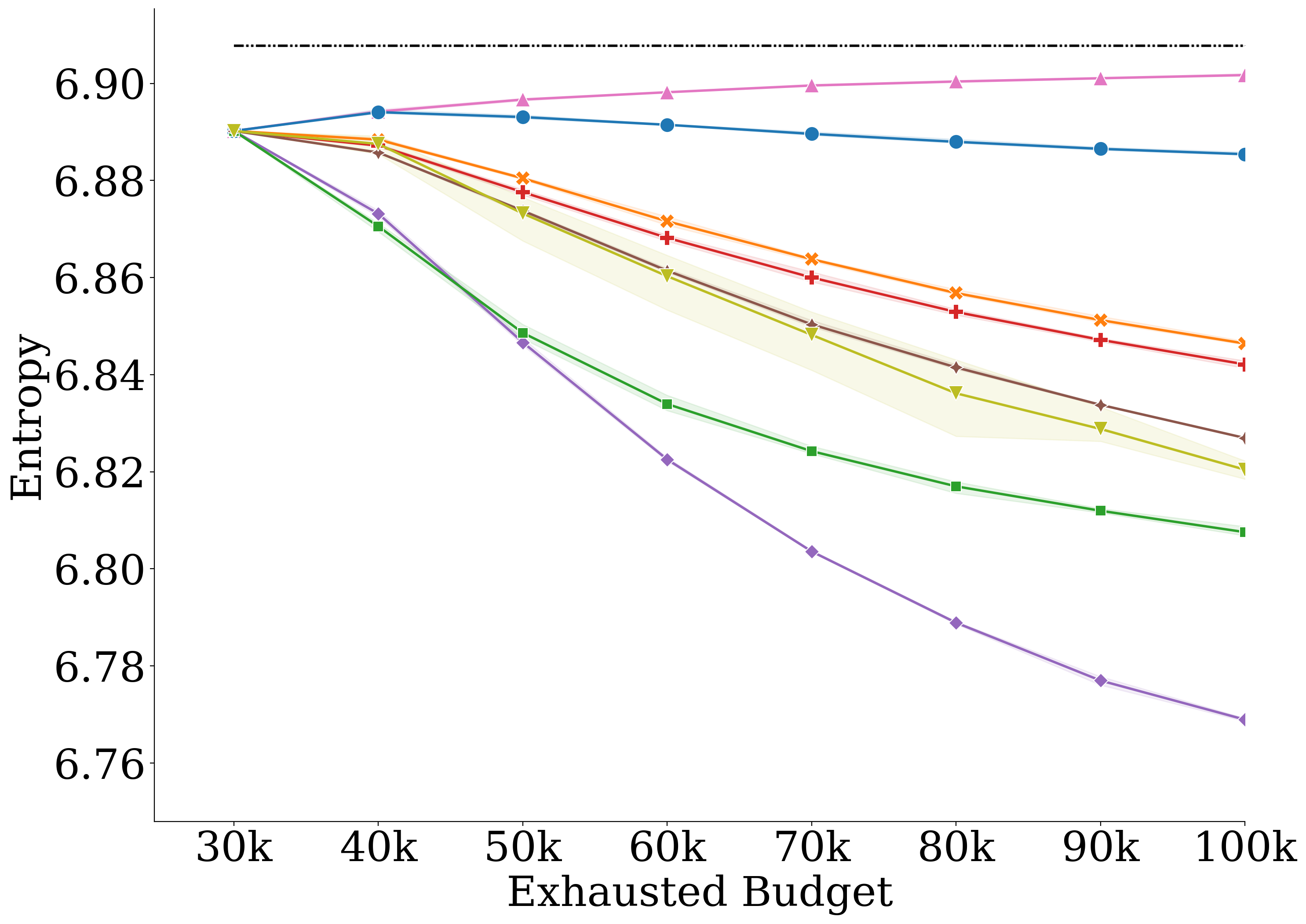}
        \caption{Setting \ref{setting:imagenet_resnet50}-\ref{setting:finetune_ssp}.}
        \label{fig:entropy_imagenet_finetune}
    \end{subfigure}
    \bigskip
    \begin{subfigure}{.48\linewidth}
        \centering
        \includegraphics[width=.98\linewidth]{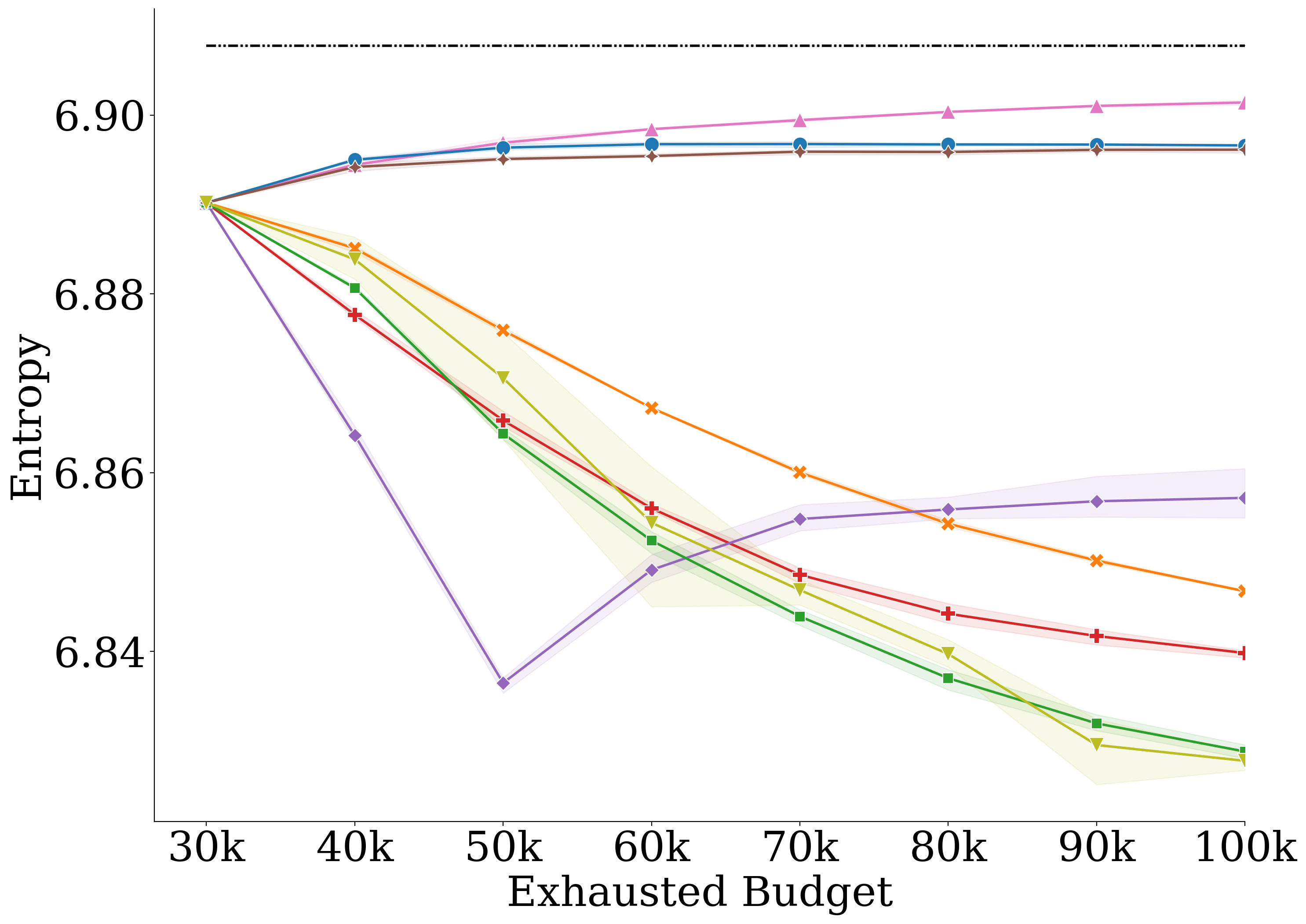}
        \caption{Setting \ref{setting:imagenet_resnet50}-\ref{setting:linear_eval}. }
        \label{fig:entropy_imagenet_linear}
    \end{subfigure}
    \hfill
    \begin{subfigure}{.48\linewidth}
        \centering
        \includegraphics[width=.98\linewidth]{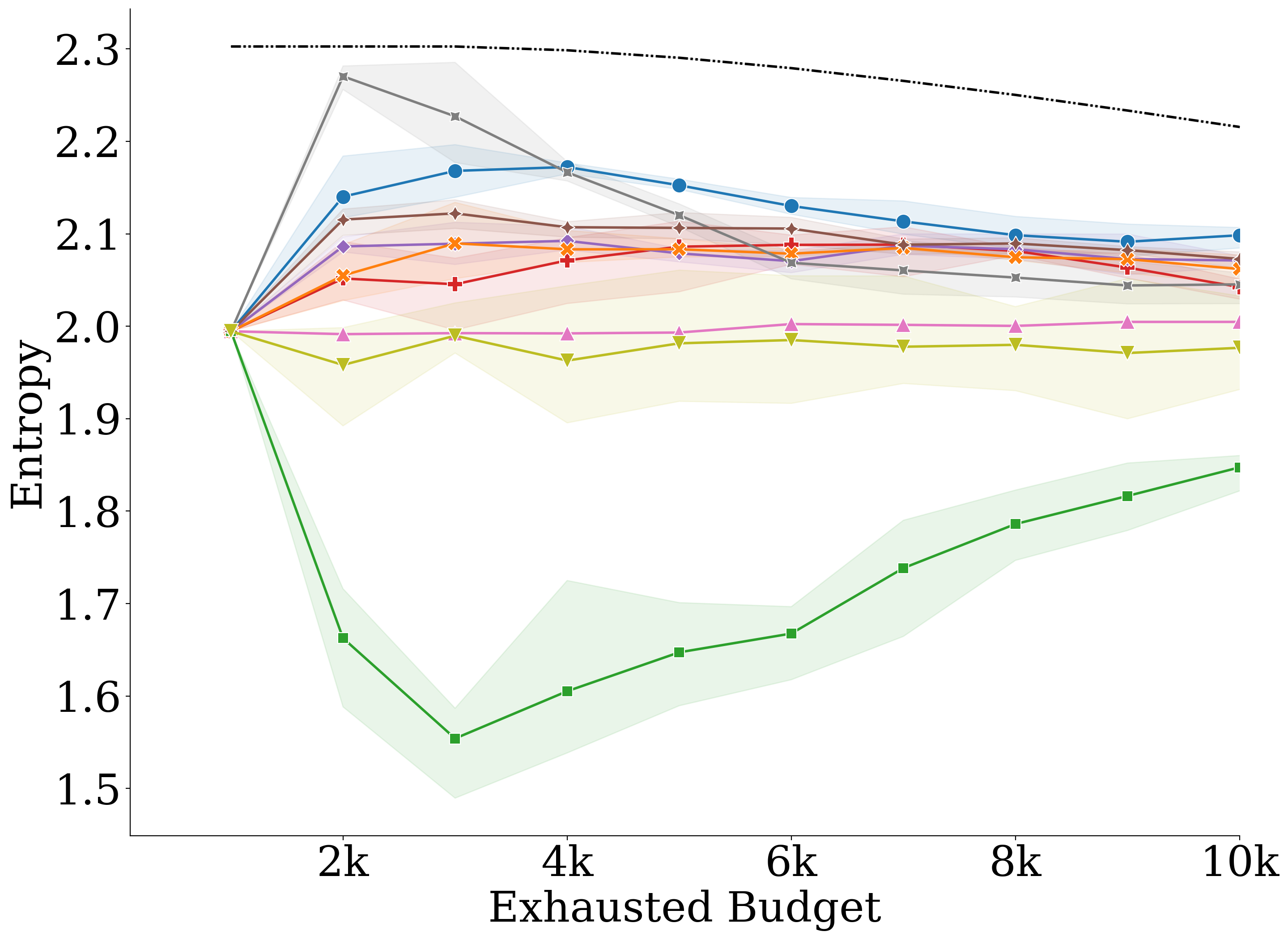}
        \caption{Setting \ref{setting:imb_cifar10_resnet18}-\ref{setting:finetune_ssp}. }
        \label{fig:entropy_imb0.1cifar10_finetune}
    \end{subfigure}
    \caption{Class distribution entropy of $D_L^k$ at different active learning rounds $k$.}
    \label{fig:entropy-plots}
\end{figure*}

Figure \ref{fig:entropy-plots}  displays the entropy of the class distributions at every AL round. Higher entropy is desirable as it indicates more balanced sampling.

\section{ImageNet Class Distribution Histograms for end-to-end finetuning}
\begin{figure*}[!t]
        \centering
        \begin{subfigure}{.01\linewidth}
            \centering
            \includegraphics[height=14.85\linewidth ]{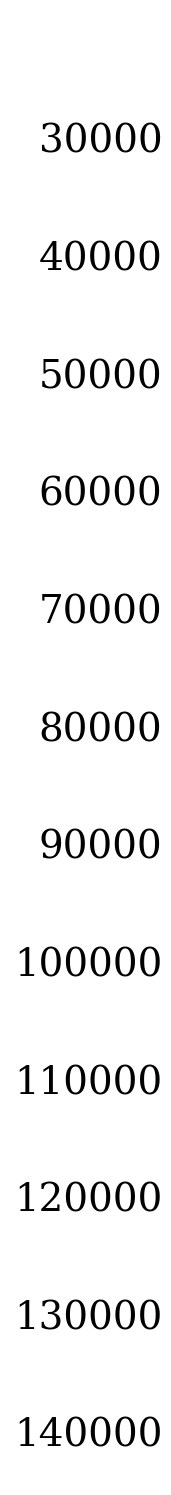}
            \caption*{\phantom{hi}}
        \end{subfigure}
        \hspace{.2mm}
        \begin{subfigure}{.11\linewidth}
            \centering
            \includegraphics[width=0.9\linewidth]{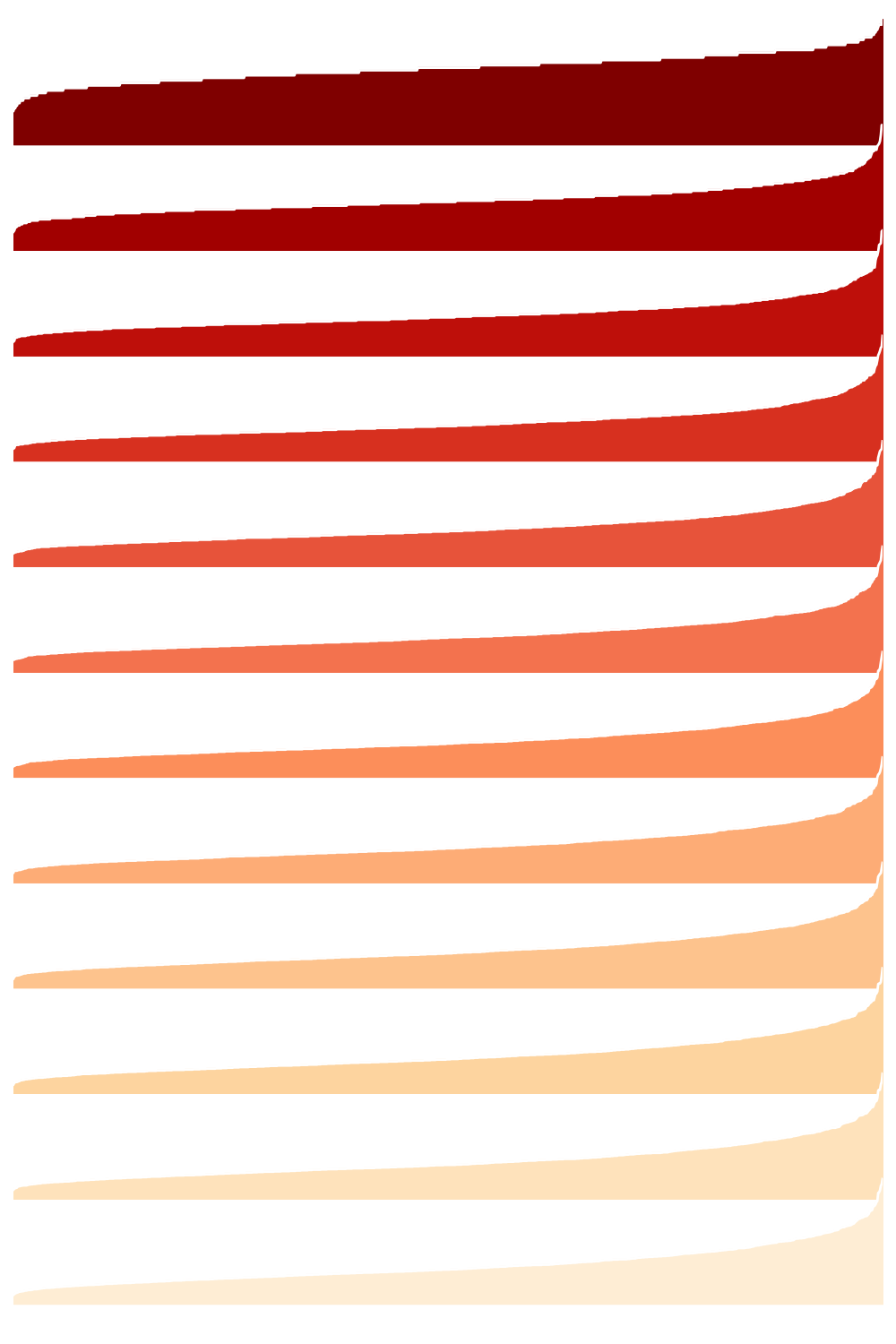}
            \caption{\ref{baseline:PartitionedCoreset}}
        \end{subfigure}
            \hfill
        \begin{subfigure}{.11\linewidth}
            \centering
            \includegraphics[width=0.9\linewidth]{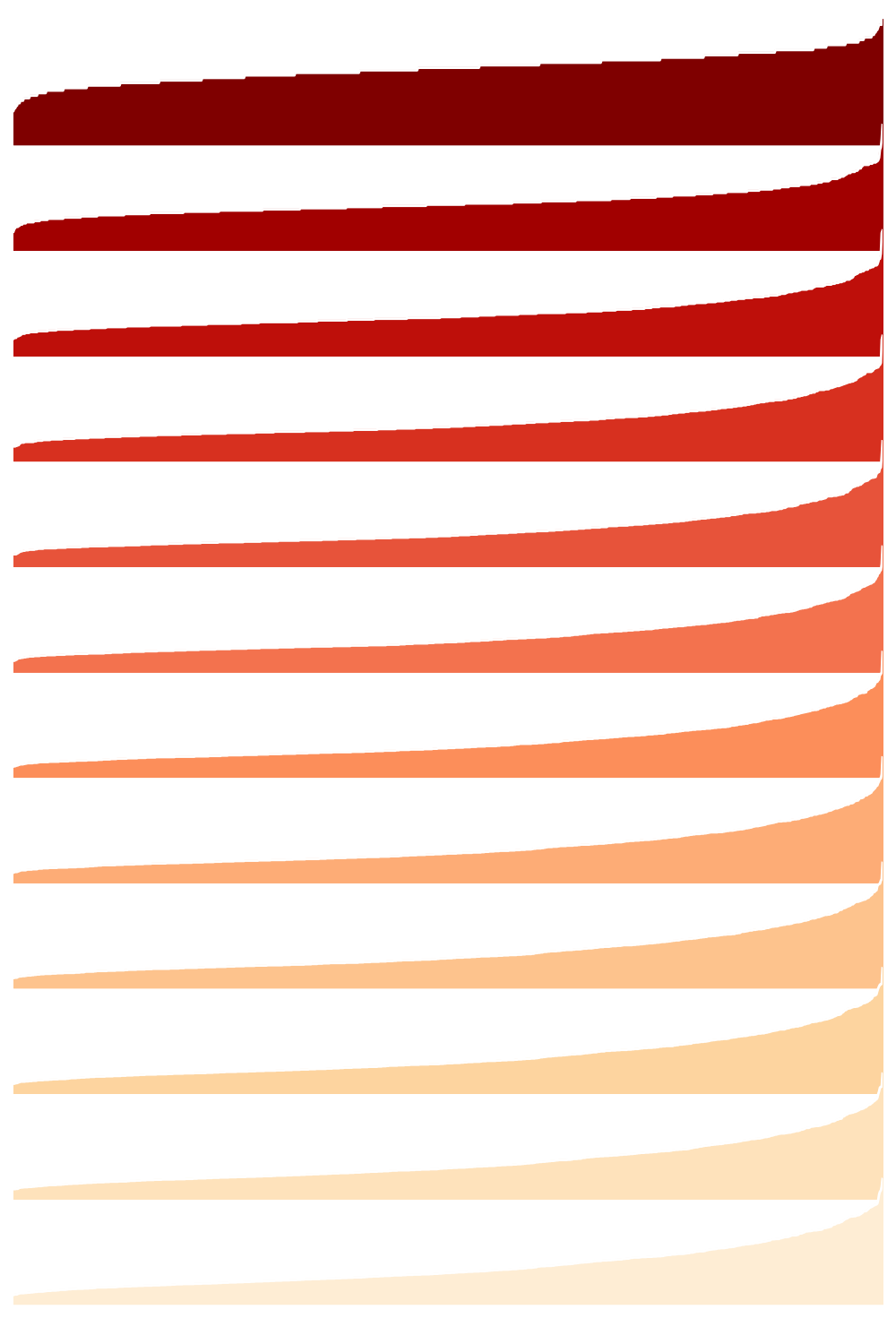}
            \caption{\ref{baseline:ConfidenceSampler}}
        \end{subfigure}
         \hfill
        \begin{subfigure}{.11\linewidth}
            \centering
            \includegraphics[width=0.9\linewidth]{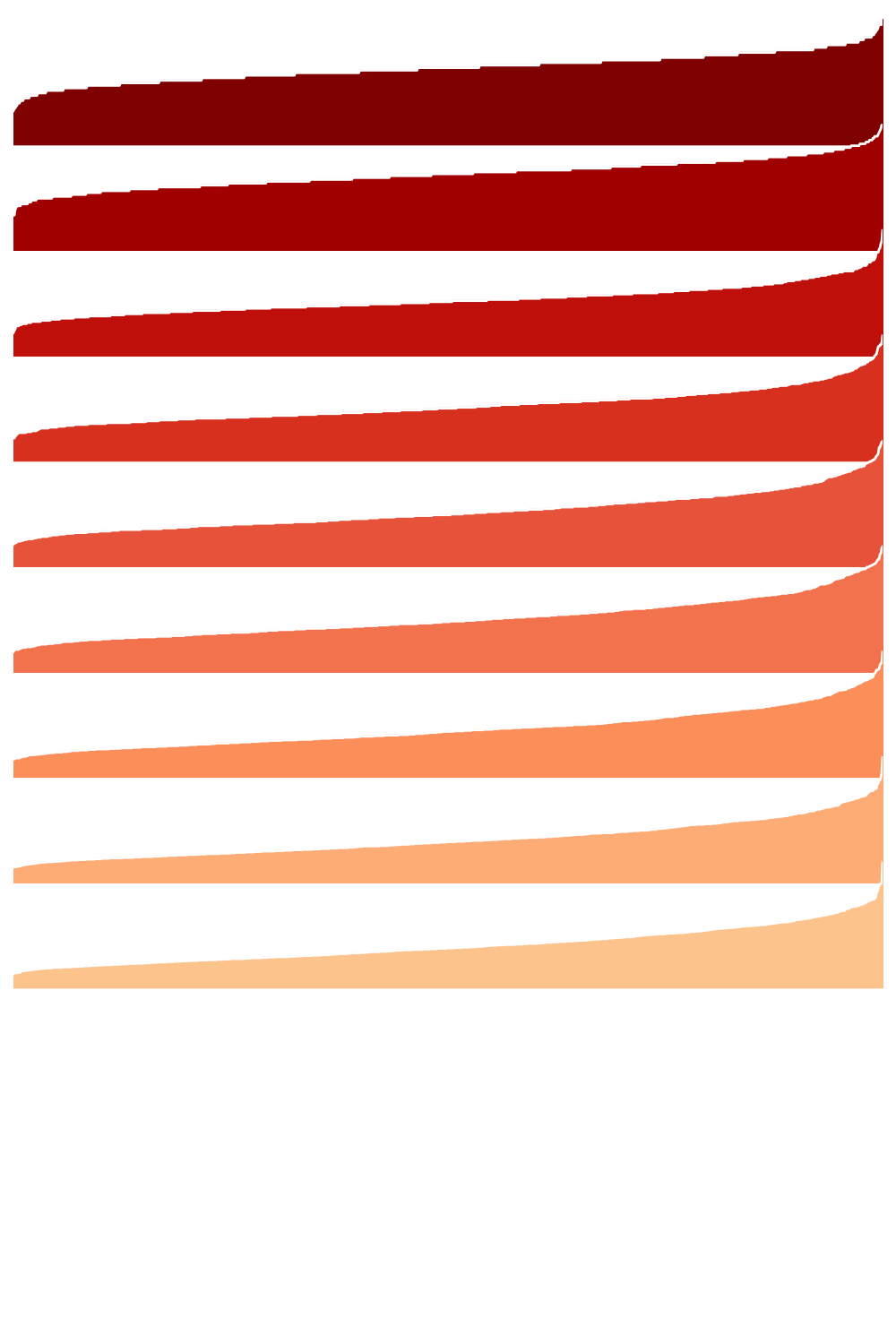}
            \caption{\ref{baseline:VAAL}}
        \end{subfigure}
         \hfill
        \begin{subfigure}{.11\linewidth}
            \centering
            \includegraphics[width=0.9\linewidth]{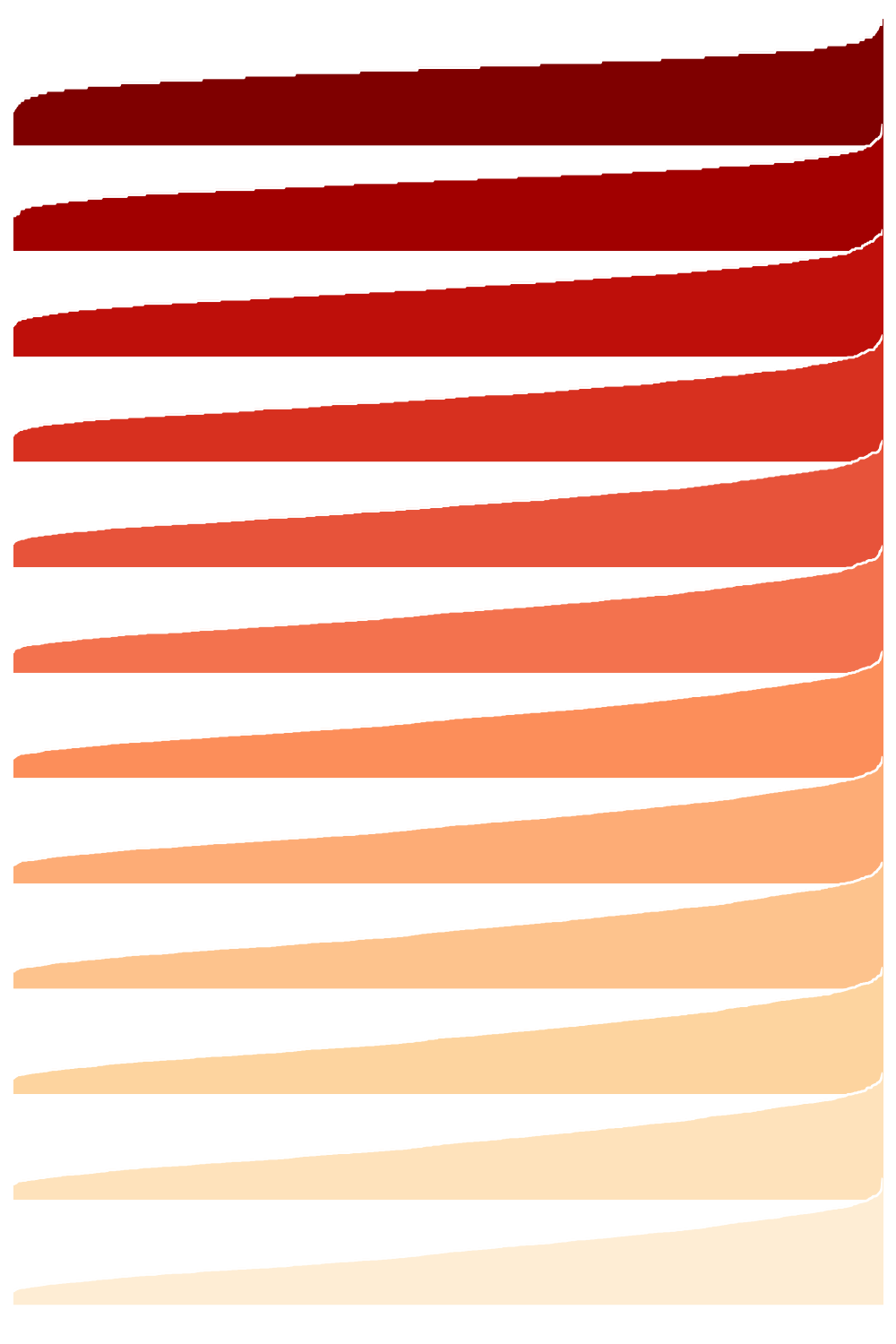}
            \caption{\ref{baseline:PartitionedBADGE}}
        \end{subfigure}
            \hfill
        \begin{subfigure}{.11\linewidth}
          \centering
          \includegraphics[width=0.9\linewidth]{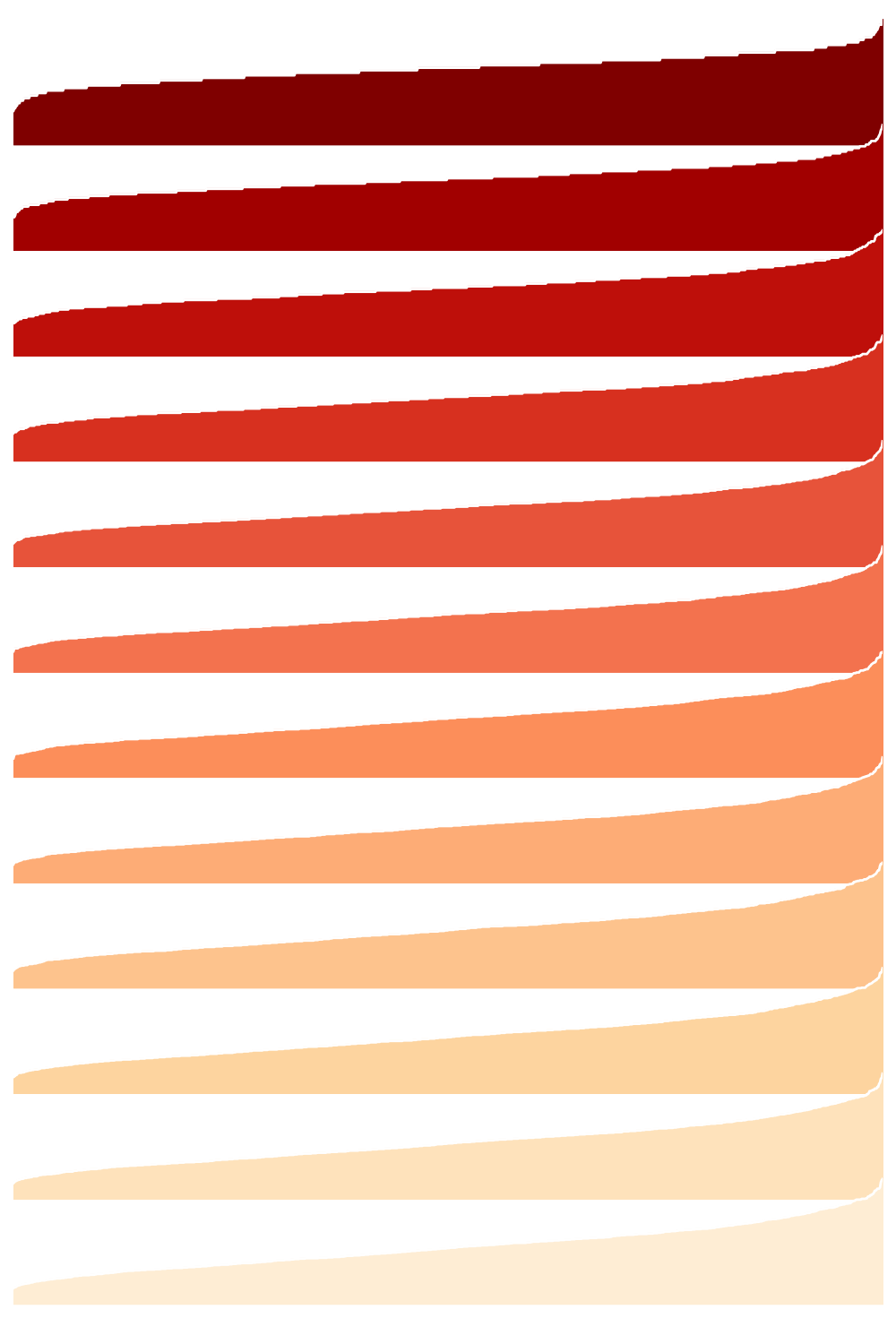}
          \caption{\ref{baseline:MarginSampler}}
        \end{subfigure} 
           \hfill
        \begin{subfigure}{.11\linewidth}
            \centering
            \includegraphics[width=0.9\linewidth]{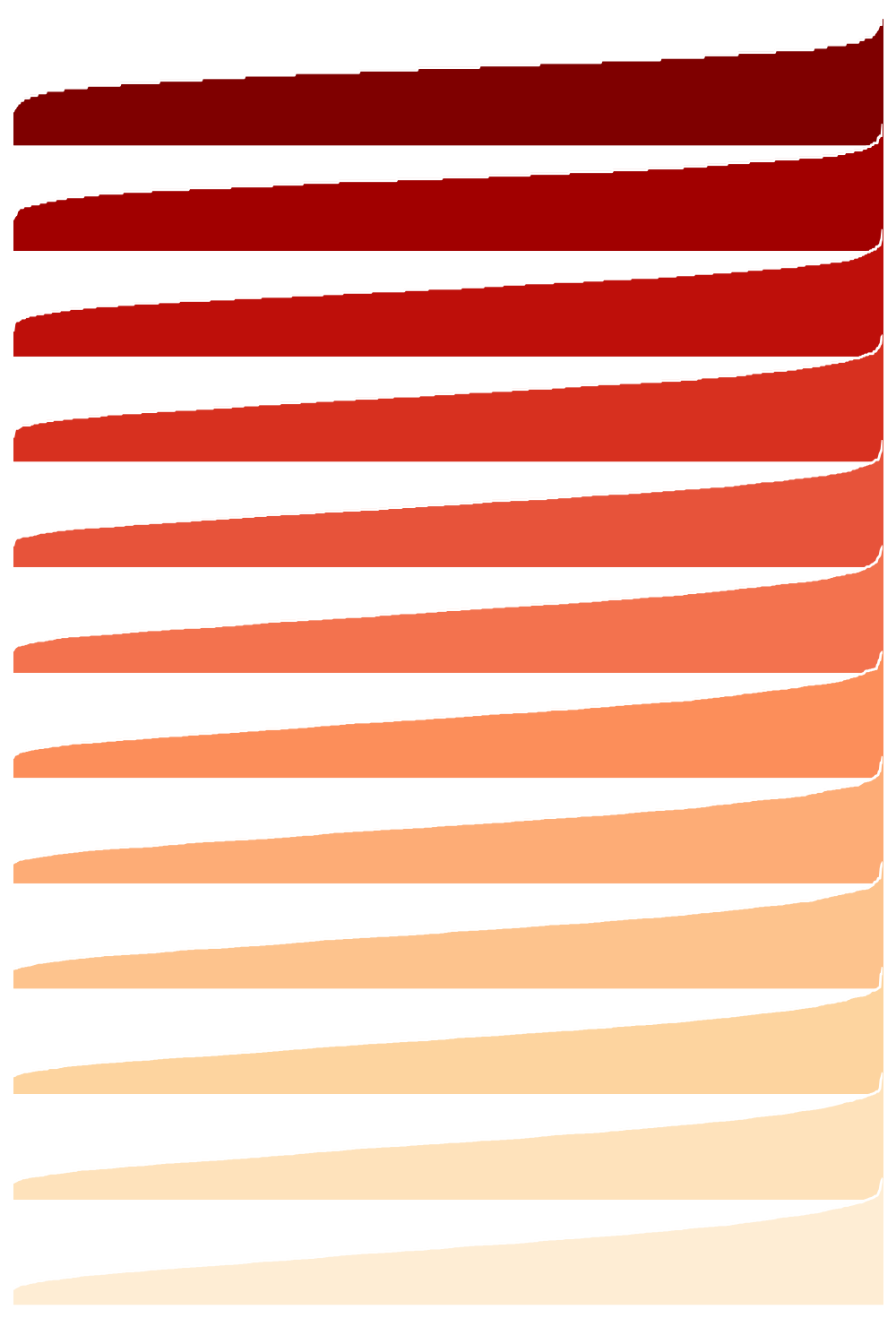}
            \caption{MASE \ref{baseline:MarginDistanceSampler}}
        \end{subfigure}
            \hfill
        \begin{subfigure}{.11\linewidth}
            \centering
            \includegraphics[width=0.9\linewidth]{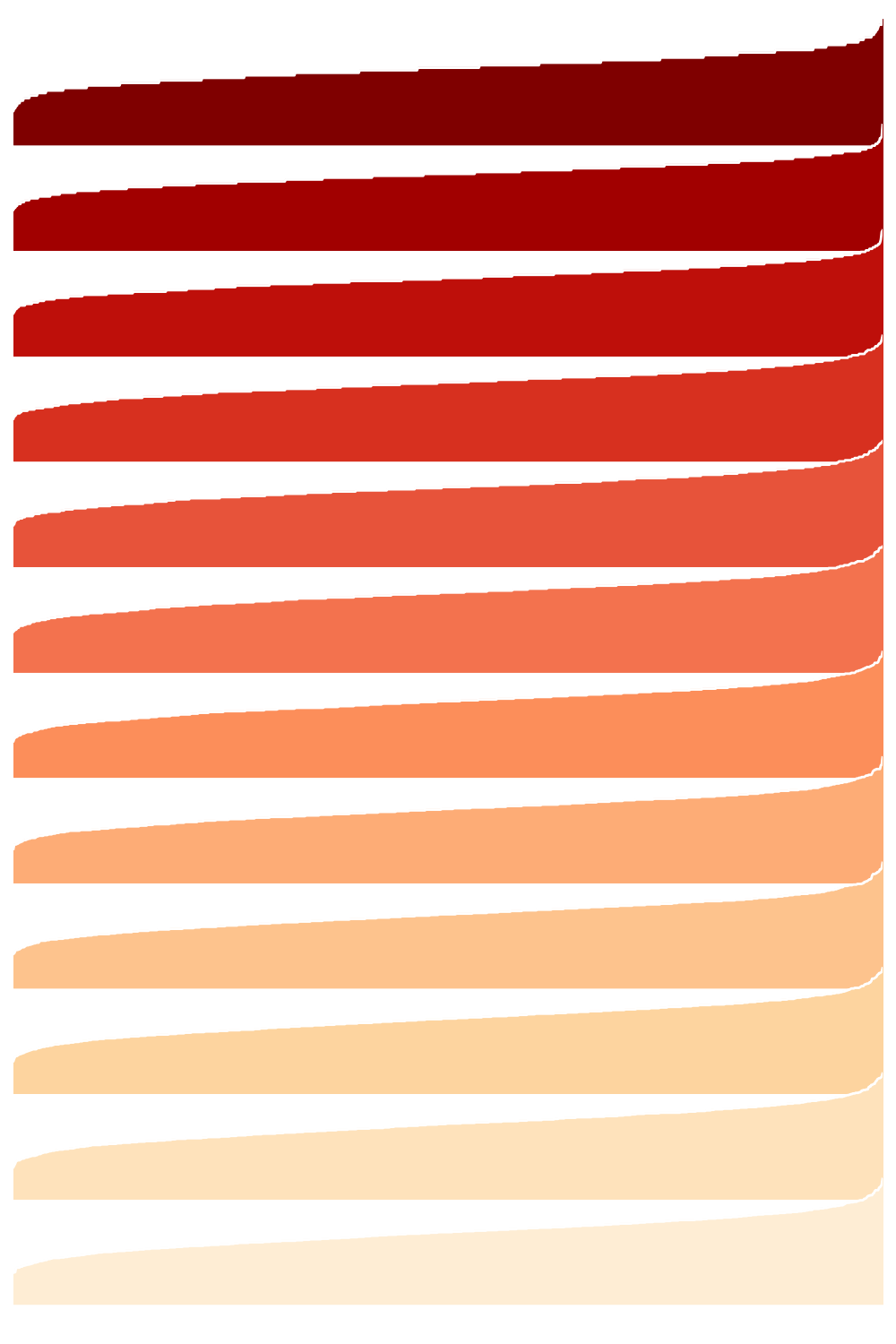}
            \caption{BASE (ours)}
        \end{subfigure}
            \hfill
        \begin{subfigure}{.11\linewidth}
            \centering
            \includegraphics[width=0.9\linewidth]{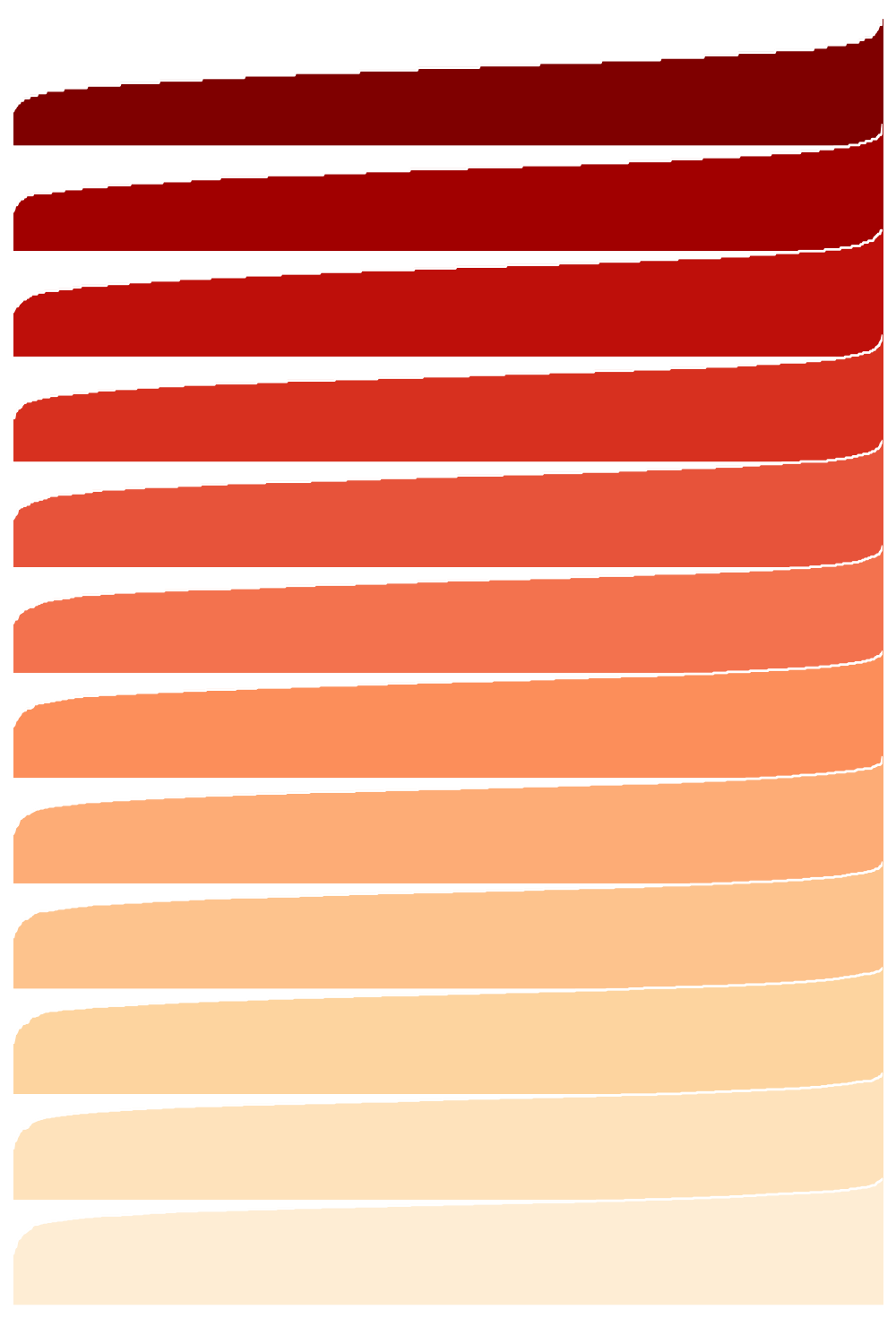}
            \caption{\ref{baseline:RandomSampler}}
        \end{subfigure}
        \caption{Setting \ref{setting:imagenet_resnet50}-\ref{setting:finetune_ssp}. The distribution of $D_U^k$ at every AL round for different strategies on ImageNet in the end-to-end finetuning setting. All experiments start with the same randomly selected subset $s^0$. The x-axis is sorted for each histogram (every row in every subplot) from least queried class to most queried class. The height of the histogram at a given location on the x-axis indicates the proportion of the examples sampled from that class. BASE is visibly the most balanced strategy after random sampling.}
        \label{fig:imagenet_finetune_histograms}
    \end{figure*}

Figure \ref{fig:imagenet_finetune_histograms} contains histograms of the distributions for our end-to-end finetuning ImageNet experiments.

%-------------------------------------------------------------------------
\section{Limitations }
    
    \label{sec:limitations}
    Plotting the test accuracy as a function of exhausted budget is common practice in AL research. However, these experiments are difficult to produce correctly as they are computationally expensive and very sensitive to hyperparameters \cite{Anonymous2022best, Beck2021effective}. We conduct thorough experiments over an extensive set of hyperparameters to ensure fair comparisons. We summarize our findings below. 
    \begin{enumerate}
        \item \textbf{Training hyperparameters at every round.} This includes the choice of optimizer, learning rate, regularization, and early stopping hyperparameters. It is necessary to train the network to saturation at every round with varying amounts of training data; otherwise, a fair comparison of AL algorithms would not be possible. In fact, if the network is not trained to saturation at a round k, the AL algorithm will not query an optimal set $s^k$, which will in turn affect the distribution of $D^{k+i}_L$ for all subsequent rounds $k+i$, $i>1$ \cite{Anonymous2022best}.
        \item \textbf{Initial Budget.} $s^0$ is randomly selected, therefore if the dataset is class balanced, $s^0$ will be relatively balanced. If $s^0$ is large, it will take many rounds of querying before we can notice performance differences between AL algorithms that select balanced data and those that don't. It is therefore important to monitor the distribution of $D_L^k$ along with the accuracy of the model at each round before drawing conclusions about performance. 
    \end{enumerate}
\newpage

%%%%%%%% Table Exp detail
    \begin{table*}
        \centering
        \begin{tabular}{c|c|c|c|c|c|c|c|c|c}
            Settings & $|s^0|$ & $b$ & Epochs & ESP & Batch size & Optimizer & Learning rate & Weight Decay & Momentum \\
            \hline
            \ref{setting:cifar10_resnet18}-\ref{setting:finetune_ssp}  & 1000  & 1000  & 200 & 50 & 128 & SGD & $1e^{-3}$ & $5e^{-4}$ & 0.9\\
            \ref{setting:imb_cifar10_resnet18}-\ref{setting:finetune_ssp}  & 1000  & 1000  & 200 & 50 & 128 & SGD & $2e^{-3}$ & 0         & 0.9\\
            \ref{setting:imagenet_resnet50}-\ref{setting:finetune_ssp}  & 30000 & 10000 & 60  & 30 & 128 & SGD & $1e^{-3}$ & 0         & 0.9\\
            \ref{setting:imagenet_resnet50}-\ref{setting:linear_eval} & 30000 & 10000 & 60  & 30 & 128 & SGD & $15$      & $1e^{-4}$ & 0.9\\
        \end{tabular}
        \caption{Hyperparameters used for each setting. $|s^{0}|$ denotes the initial pool size. $b$ denotes the budget per round, and ESP abbreviates early stop patience.}
        \label{tab:hyper_param}
    \end{table*}
%%%%%%%   

%-------------------- ENTROPY Plots

%--------------------------- END-TO-END FINETUNING IMAGENET HISTOGRAMS

\end{document}

%% file: shortcuts.tex
\newcommand{\EE}{{\mathbb E}}

\newcommand{\RR}{{\mathbb R}}

\newcommand{\Xcal}{\mathcal{X}}
\newcommand{\Ycal}{\mathcal{Y}}
\newcommand{\Zcal}{\mathcal{Z}}

\def\benm{\begin{enumerate}}            % Begin enumerate command
\def\eenm{\end{enumerate}}              % End enumerate command
\DeclareMathOperator*{\argmin}{arg\,min}